\documentclass[twocolumn,letterpaper]{IEEEAerospaceCLS}  

\usepackage[]{graphicx}    
\usepackage[]{subcaption}
\usepackage[]{tabularx}
\usepackage[]{amsmath}
\PassOptionsToPackage{hyphens}{url}\usepackage{hyperref}
\newcommand{\ignore}[1]{}  

\begin{document}
\title{The Landform Contextual Mesh: Automatically Fusing Surface and Orbital Terrain for Mars 2020}


\author{%
Marsette Vona\\
Jet Propulsion Laboratory, California Institute of Technology\\
4800 Oak Grove Dr.\\
Pasadena, CA 91011\\
marsette.a.vona@jpl.nasa.gov
\thanks{\footnotesize \copyright2025 California Institute of Technology.  Government sponsorship acknowledged.}              
}

\maketitle

\thispagestyle{plain}
\pagestyle{plain}

\maketitle

\thispagestyle{plain}
\pagestyle{plain}

\begin{abstract}
The Landform contextual mesh fuses 2D and 3D data from up to thousands of Mars 2020 rover images, along with orbital elevation and color maps from Mars Reconnaissance Orbiter, into an interactive 3D terrain visualization. Contextual meshes are built automatically for each rover location during mission ground data system processing, and are made available to mission scientists for tactical and strategic planning in the Advanced Science Targeting Tool for Robotic Operations (ASTTRO).  A subset of them are also deployed to the "Explore with Perseverance" public access website.
\end{abstract} 

\tableofcontents

\section{Introduction}
Like its predecessors including Spirit, Opportunity, and Curiosity, the Mars 2020 Perseverance rover carries a suite of stereo cameras to image the surrounding terrain~\cite{farley2020mars,maki2020mars,M20CamSIS}.  Data from those instruments is used for multiple purposes including both on-board and ground-based navigation, engineering operations, and science analysis.  An important use case is tactical and strategic science planning, where teams of mission scientists use the imagery to select areas of interest and plan subsequent observations.

\begin{figure*}
\centering
\begin{subfigure}{0.331\textwidth}
    \includegraphics[width=\textwidth]{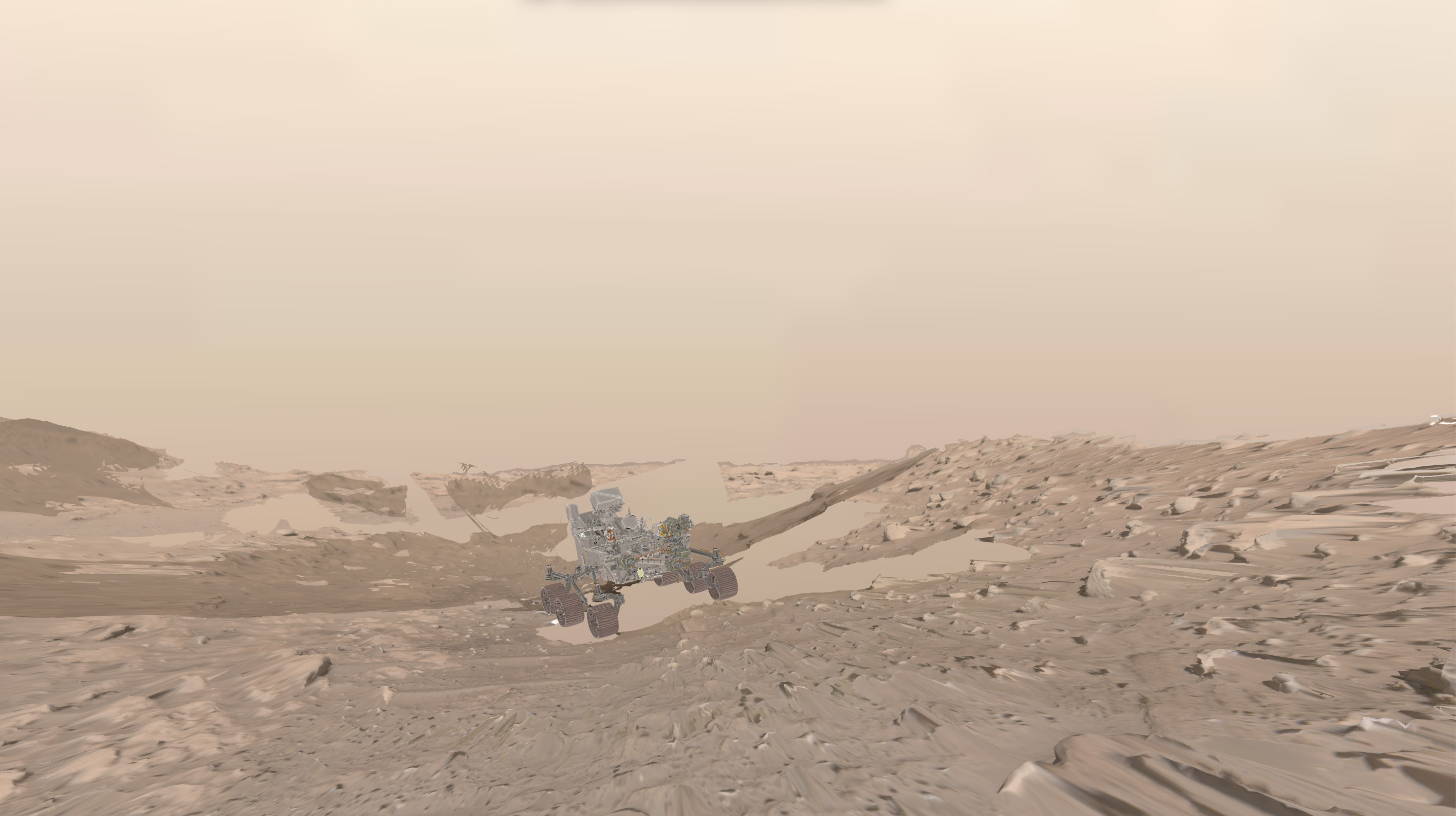}
    \caption{Tactical mesh from in-situ stereo vision only.}
    \label{SubfigTactical}
\end{subfigure}
\hfill
\begin{subfigure}{0.328\textwidth}
    \includegraphics[width=\textwidth]{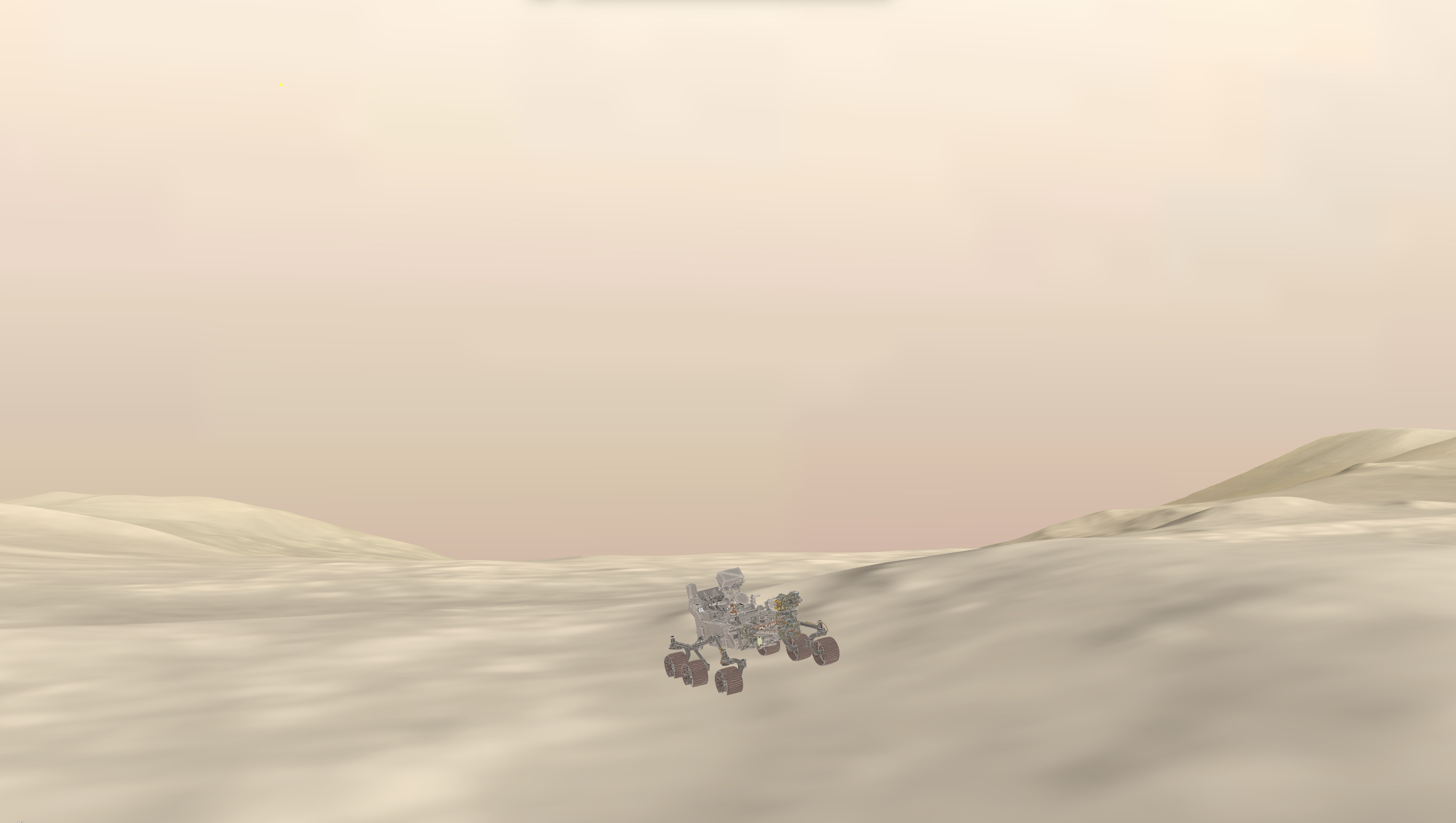}
    \caption{Orbital digital elevation mesh from Mars Reconnaissance Orbiter HiRISE data.}
    \label{SubfigOrbital}
\end{subfigure}
\hfill
\begin{subfigure}{0.331\textwidth}
    \includegraphics[width=\textwidth]{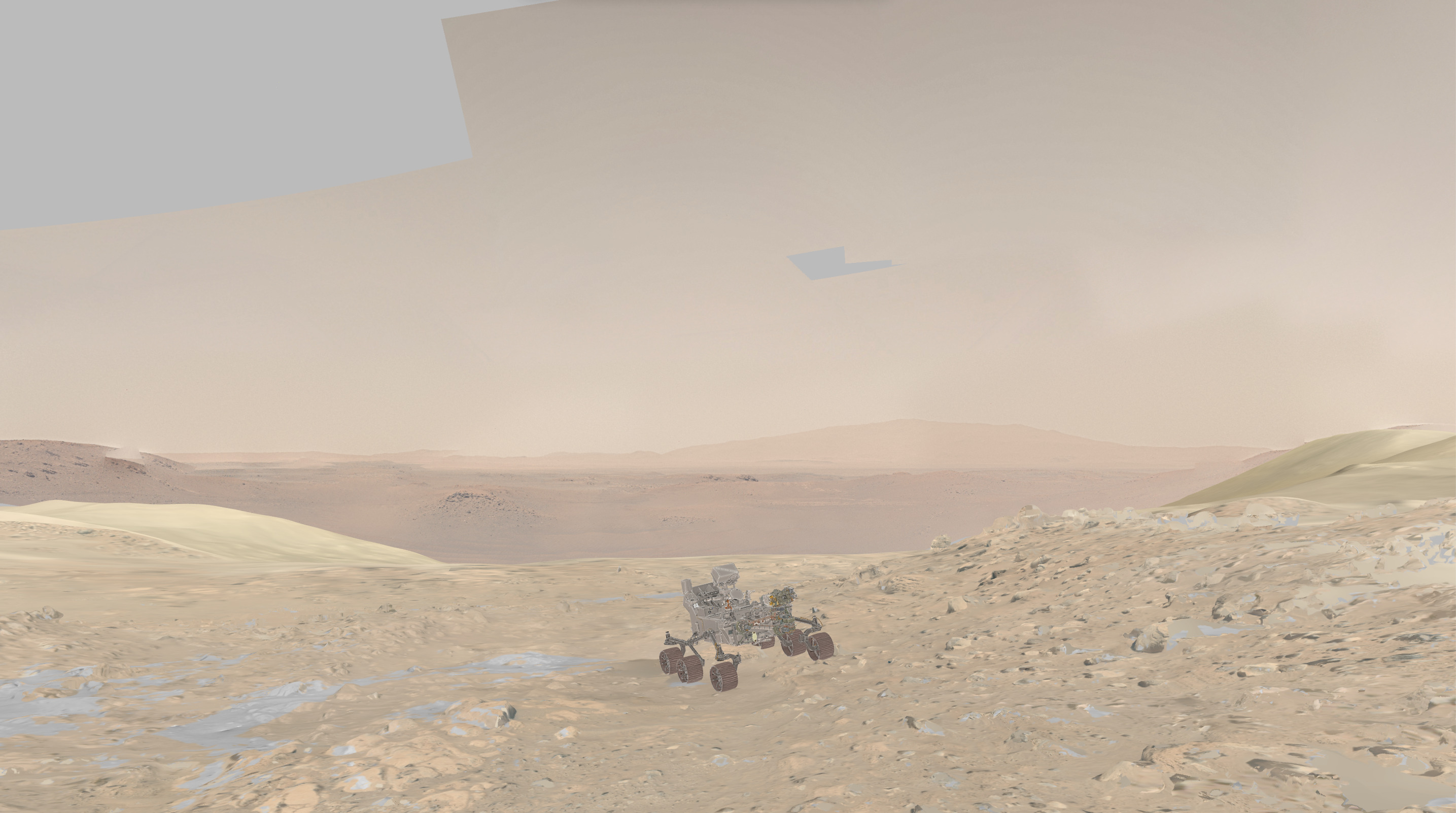}
    \caption{Landform contextual mesh fusing in-situ and orbital data.}
    \label{SubfigContextual}
\end{subfigure}
\caption{Reconstructed 3D terrain surrounding the Perseverance rover at site 40, drive 132 (sols 821--832).  This viewpoint spans about 15m horizontally in the foreground and about 1km from foreground to background terrain.  More distant hills are also visible in the sky sphere.}
\label{FigTacticalOrbitalContextual}
\end{figure*}

Stereo vision produces a ``tactical wedge'' 3D terrain mesh for each stereo image pair, so-called because often a radial panorama of such wedges is acquired from a single rover location using the pan/tilt mast.  Of course, only a fraction of nearby terrain is included since the cameras have limited fields of view, effective resolution decreases with distance, the rover occludes areas underneath itself, nearby rocks and hills create self-occlusions in the terrain, and stereo reconstruction fails in areas with insufficient texture.  Nevertheless, viewing such panoramas of tactical wedges in 2D and 3D (latter also called the \emph{tactical mesh}) has been a standard approach for science planning on Mars 2020 and its predecessors.  Figure~\ref{SubfigTactical} shows such a view for site 40, drive 132 of the Perseverance rover, acquired on sols 821--832\footnote{One \emph{sol} is the equivalent of a day on Mars.}.

An alternative to the tactical mesh is to use a portion of a digital elevation map (DEM) derived from orbital observations, as shown in Figure~\ref{SubfigOrbital}.  The Mars 2020 mission typically uses colored DEM data from the HiRISE instrument on the Mars Reconnaissance Orbiter~\cite{McEwenHiRISE}.  This \emph{orbital mesh} can cover a much larger extent---up to 10s of km---and typically has no gaps in its coverage.  However, at a typical resolution of 1 elevation sample (and 16 color samples) per square meter, it's much coarser than the tactical mesh, which can have sub-millimeter resolution near the rover.

\begin{figure*}
\centering
\includegraphics[width=\textwidth]{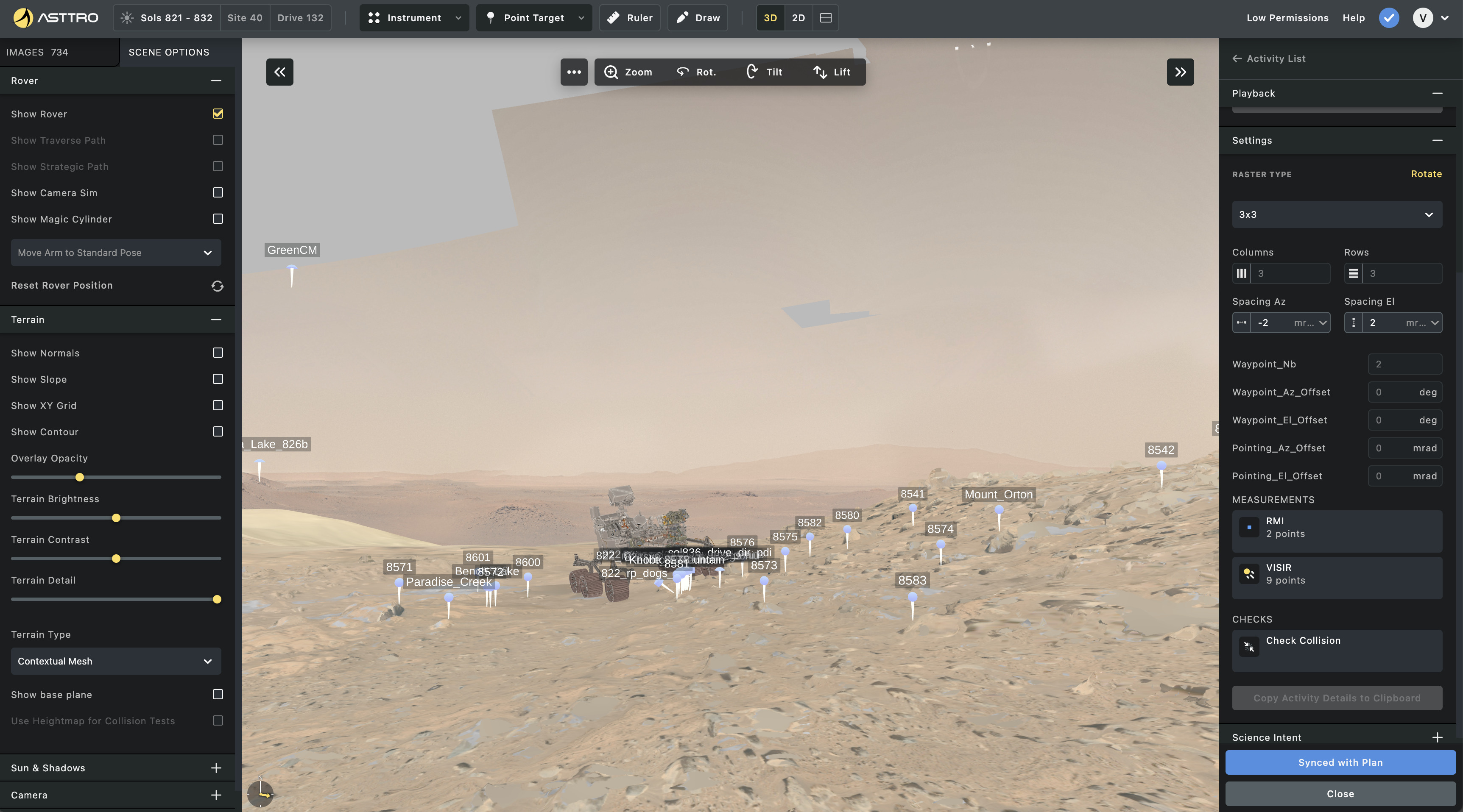}
\caption{Landform contextual mesh viewed in the Mars 2020 ASTTRO tool for science planning (same viewpoint as Figure~\ref{FigTacticalOrbitalContextual}).  Named targets are typically defined on the tactical mesh for accuracy but can be viewed relative to the contextual mesh for additional spatial awareness.}
\label{FigASTTRO}
\end{figure*}

In this paper we introduce the \emph{contextual mesh}, which we have developed to fuse up to thousands of images from in-situ stereo cameras together with orbital DEM data into a single 3D scene, shown in Figure~\ref{SubfigContextual} and Figure~\ref{FigASTTRO}.  The contextual mesh is produced by Landform, a subsystem within the Mars 2020 ground data system (M20 GDS), and typically viewed in the Advanced Science Targeting Tool for Robotic Operations (ASTTRO) collaborative web application~\cite{Paar2023,Abercrombie2019}, also part of the M20 GDS~\cite{Pyrzak2022}.

Whereas the tactical mesh offers the highest fidelity local terrain reconstruction, and the orbital mesh the longest range reconstruction, the intention of the contextual mesh is to provide spatial awareness. It is typically visualized from a first-person navigable 3D point of view in ASTTRO, showing not only local terrain features such as sand, pebbles, rocks, ridges, and hills, but also distant landmarks on the horizon.  ASTTRO also displays a 3D model of the Perseverance rover on the terrain as it was posed at the corresponding time in the mission.

Each contextual mesh is comprised of two tilesets in the open-standard 3DTiles format~\cite{Cesium3DTiles}.  One tileset contains the terrain itself, typically extending to a 1km square with a 100m square central detail area.  The other tileset is a hemispherical representation of the surrounding horizon and sky, enabling visualization of distant features potentially many kilometers away.  ASTTRO displays both of these simultaneously so that users can see the context of local terrain features relative to both nearby hills and to the horizon.

The 3DTiles format enables data to be progressively streamed to distributed users in the web-based ASTTRO client.  Only the subset of tile data required depending on the user's current viewpoint is transferred and rendered, enabling dynamic level-of-detail, fast load times, and deployment to resource limited clients.

Like the products of many data fusion and reconstruction algorithms, e.g. computed tomography, and considering that the input data contains noise and outliers, the contextual mesh may contain some artifacts. For example
\begin{itemize}
\item boundaries between areas reconstructed primarily from surface vs orbital data may have some discontinuities
\item outlier images with extreme brightness variations may not be completely blended
\item reconstructed geometry may have ``island'' topological artifacts due to noise and residual misalignment in the input data.
\end{itemize}
Spatial awareness can generally still be gained even in the presence of such artifacts.  The tactical or orbital mesh can be used for other tasks, and ASTTRO makes it easy to change to the tactical, orbital, or contextual mesh at any time.

The Landform contextual mesh differs from many other photogrammetry and terrain fusion systems not only because it combines both surface and orbital data and has a sky sphere, but also in that it is entirely automated, whereas most other systems require some human intervention.  It heavily leverages properties of Mars mission datasets, including pose priors from rover navigation, calibrated stereo camera data, and co-registered orbital data. These enable automated processing while maintaining reasonable quality relative to manual and semi-manual approaches.

The Landform codebase will soon be released as open source.  And, throughout the mission, a selection of contextual meshes have been made publicly available for interactive viewing at the ``Explore with Perseverance'' website~\cite{ExploreWithPerseverance}.

In this paper we summarize the research context of related terrain fusion approaches, describe the novel algorithms we developed to implement the contextual mesh system in Landform, and present examples of contextual mesh data products.

\section{Related Work}
The Landform contextual mesh is effectively a stereo-vision based simultaneous localization and mapping (SLAM) system which operates on data from stereo cameras\footnote{Even the orbital DEM data is based on stereo vision processing of image pairs from the MRO HiRISE camera~\cite{McEwenHiRISE}.} and which produces a dense reconstruction of the 3D scene.  While early SLAM systems were often based on sparse feature representations~\cite{Dissanayake01,Montemerlo02,SLAMI,SLAMII,PTAM}, many later systems are dense~\cite{DTAM,Frahm10,Geiger11,Engel15}.  Landform uses sparse features for alignment, as detailed below, but still produces a dense reconstruction.

There is wide variation in the types of input sensors used in various SLAM systems in the literature, including monocular cameras~\cite{PTAM}, depth cameras~\cite{Newcombe11}, lidar~\cite{Hess16}, and stereo vision~\cite{Geiger11,Engel15,ORBSLAM2}.  However, few previous systems attempt to combine data from both in-situ surface stereo imagery with stereo reconstructions from remotely sensed orbital images.  Also, the well-calibrated cameras and kinematics of the rover enable Landform to operate on much coarser chunks of input data than many other SLAM systems.  Whereas generic stereo SLAM systems typically consider each stereo pair as a separate entity to localize, Landform instead operates on entire groups of data from each rover location.  Landform also leverages the mission's good (though still imperfect) estimates of rover poses across drive segments and relative to the orbital data to seed the localization process, reducing the need for explicit global optimization.

Many other SLAM systems are also formulated as on-line incremental algorithms which work in real time, appropriate for active use in robot locomotion.  As a ground-based system, Landform operates on a large time delay relative to the rover, partly due to the light time from Mars to Earth, but mostly dominated by latencies in scheduling communications via the deep space network (DSN).  The Landform contextual mesh is thus an off-line system related to other work in scene reconstruction for teleoperation~\cite{Dejing18,Jin21} and 3D map-making~\cite{Frahm10,Geiger11,Engel15}.

The Landform contextual mesh can also be compared to 3D scene reconstructions made using a variety of commercial photogrammetry packages~\cite{RealityCapture,Agisoft,Pix4D}.  Indeed, given sufficient time---possibly measured in days---a skilled human operator using the same input data as Landform could often produce better results using such packages.  However, the Landform contextual mesh addresses the question of what can be done by a fully automated unattended system in a relatively short amount of time---contextual meshes are usually produced automatically within 4--8 hours after the input data appears.

\section{Ground Data System Interface}
The contextual mesh is a synthetic data product produced during ground data processing for Mars 2020.  The inputs for the contextual mesh are:
\begin{itemize}
\item VICAR~\cite{VICARFormat} format radiometrically corrected ``RAS'' reduced data record (RDR) images~\cite{M20CamSIS} from the Hazcam, Navcam, and Mastcam-Z stereo cameras
\item per-wedge point cloud ``XYZ'' RDRs, the result of stereo vision processing for each stereo pair, formatted as 3 band floating point VICAR images, along with corresponding ``UVW'' surface normal RDRs
\item rover mask ``MXY'' RDRs, formatted as single band binary VICAR images, flagging areas of the other images expected to contain the rover mechanism itself vs natural terrain
\item VICAR header metadata for each RDR including CAHVORE~\cite{Gennery2006} intrinsic and extrinsic camera model parameters, rover motion counter (RMC)~\cite{M20PlacesSIS}, rover joint angles, image color model, and framing metadata
\item rover pose estimates from the ``PLACES'' database~\cite{M20PlacesSIS}
\item MRO HiRISE orbital DEM as a single band floating point GeoTIFF~\cite{GeoTIFFFormat}, 1 pixel per meter
\item MRO HiRISE color orbital orthoimage as a three band RGB GeoTIFF, 4 pixels per meter.
\end{itemize}
All but the last two are produced from raw downlink data by earlier parts of the M20 instrument data system (IDS) pipeline.  The orbital GeoTIFF images cover a broad area of M20 rover operations and are infrequently updated.

The specific set of rover image RDRs used for the contextual mesh is configurable, and in practice there are a variety of options and variants to consider.  The Mars 2020 camera software interchange specification (SIS) gives details~\cite{M20CamSIS}.  Typically we only use data from the two mast-mounted stereo cameras---Navcam and Mastcam-Z---as well as the front and rear-facing stereo Hazcams.  These instruments alone produce a variety of image RDRs at varying levels of processing.  We typically use only nonlinear\footnote{Nonlinear camera model intrinsics, i.e. the O, R, and E parameters in CAHVORE, have not been factored out.} images that have been radiometrically corrected to compensate for the sensor response curve.  Variations in commanding can produce images of different sizes and resolutions, for example if only a subframe was commanded, and even different colorspaces---while these instruments are each capable of full color, sometimes only grayscale images are commanded to save downlink bandwidth.

Each stopping location of the rover is defined by two integers: \emph{site} is typically a 3 digit integer for each general area of operations, starting at landing site 1; \emph{drive} is typically a 4 digit integer that increments from 0 as the rover completes each drive segment within a site.  These two numbers are the first two components of every RMC value~\cite{M20PlacesSIS}.  We combine them to form a \emph{sitedrive} that uniquely names each rover stopping location, e.g. the scene in Figures~\ref{FigTacticalOrbitalContextual} and~\ref{FigASTTRO} is at site 40, drive 132, and thus sitedrive 0400132.

Contextual meshes are always centered on a specific sitedrive, but include data from spatially nearby peripheral sitedrives, i.e. nearby locations of the rover either before or after the central sitedrive along the rover's traversal path.  This makes the contextual mesh relatively unique among Mars 2020 data products in that it combines in-situ image data from multiple rover locations into a single fused product.  Normally this is not attempted because the uncertainty in odometry-based pose estimates across rover drive positions can be much higher than the kinematics-based relative pose estimates for data products at a single rover location.  The alignment algorithms described below are one of the most important aspects of the contextual mesh because they enable fusing data across rover positions.

\subsection{Deployment Architecture and Provisioning}
The Landform contextual mesh subsystem is typically deployed as a set of servers on the Amazon Web Services (AWS) GovCloud.  There is one \emph{contextual master} instance always running on a relatively lightweight instance type (e.g. two processing cores, 4GB RAM).  The role of the contextual master is to do bookkeeping and monitoring to determine when to build a new contextual mesh for a given sol, central sitedrive, and set of peripheral sitedrives.

There is also a pool of much more powerful \emph{contextual worker} instances (e.g. 72 processing cores, 144GB RAM) which automatically activate and deactivate as necessary to actually produce each contextual mesh.  A worker typically takes around 4--8 hours per contextual mesh.

Once a contextual mesh has been computed the resulting 3DTiles tilesets---one for the terrain and one for the sky sphere---are simply directories of files.  They are saved to the mission operational data store (ODS) and are served by a standard web server to the ASTTRO clients.

At this time the contextual mesh data products are not included in planetary data system (PDS) archive deliveries~\cite{M20PDS}, and are only available internally within the Mars 2020 mission GDS and on the Explore with Perseverance public outreach website~\cite{ExploreWithPerseverance}.

\subsection{Automatic Triggering}
The contextual master receives messages when the IDS pipeline produces new RDR products for each pass of downlink data sent by the Perseverance rover via the DSN.  It uses several heuristics to determine that RDR generation has completed processing for a given DSN pass, including simply watching for relatively long pauses in the appearance of new products.  Each pass may include data for multiple sitedrives, and because data is often downlinked out of order with respect to when it was acquired by the rover, each pass may also include data for multiple sols.  The contextual master determines the highest numbered sitedrive for each sol that had new data in the pass, and potentially schedules a new contextual mesh centered at each of those locations.  The contextual master then checks ODS to find other already-existing RDR data for potentially nearby sitedrives within $\pm100$ sols of each sol with new data.  Then it queries the PLACES database, which tracks every rover pose, to check the distance of those other sitedrives to the central sitedrive of each new contextual mesh.  It keeps only those that are within a configurable threshold, typically set to 128 meters.  Other filtering is also applied including rejecting sitedrives with too little data.

The result of this is a set of messages sent from the contextual master to the worker pool describing new contextual meshes to be built.  Each message contains the following information:
\begin{itemize}
\item the central sitedrive, e.g. 0261222
\item the additional peripheral sitedrives to be included, e.g. 0260850, 0261004, 0261110, 0261154
\item the sol number in which the central sitedrive had new data, e.g. 534
\item sol ranges which have relevant RDR data in either the central or peripheral sitedrives, e.g. 474--475, 477--486, 488--492, 494--499, 501--534
\end{itemize}
The resulting contextual mesh will be uniquely named by combining the 4 digit sol number with the central sitedrive and an optional version number, i.e. 0534\_0261222V01.

\subsection{Ingestion and Pose Priors}
The first stage of processing a contextual mesh on a worker is to download and index the input data.  The worker checks ODS to find the paths of all candidate input RDRs in the sol ranges specified by the master.  For 0534\_0261222 this yields about 7000 potential input RDRs.  It then applies a set of rules and heuristics to trim this set, including
\begin{itemize}
\item use only the latest version of each product
\item if there are more than a configurable threshold (e.g. 400) of images for an instrument within a single sitedrive, keep only the newest within that limit
\item if there are both linear and nonlinear products, keep only the nonlinear ones 
\item if XYZ/UVW point cloud data has been computed from both the left and right eye perspective, keep the XYZ/UVW for the left eye only.
\end{itemize}
For 0534\_0261222 this reduces the set of input RDRs to about 3000 files, about 55GB of data.  The number of RDR input products of each type for this example are shown in Tables~\ref{RDRsPerProductType} and~\ref{RDRsPerSiteDrive}.  VICAR format RDRs are typically uncompressed\footnote{VICAR does support compression but it's not well documented or widely implemented.}.  Though many products are only sub-frame images, a full-frame Navcam or Hazcam is 5120x3840 or about 20 megapixels, and thus about 110MB\footnote{The radiometrically corrected ``RAS'' image RDRs used by Landform are typically three bands of 12 bit data stored in 16 bits per band.}.  Corresponding full-frame Navcam and Hazcam XYZ and UVW RDRs are about 220MB each since they have three bands with a 4 byte floating point value per band per pixel.  Mastcam-Z RDRs are about 1/10 that size since the full frame Mastcam-Z resolution is 1648x1200, or about 2 megapixels.
\begin{table}
\renewcommand{\arraystretch}{1.3}
\centering
\begin{tabular}{|r|l|l|l|l|}
\hline
              & \bf RAS & \bf XYZ & \bf UVW & \bf MXY \\
\hline\hline
\bf Hazcam    & 166     & 83      & 83      & 164     \\
\bf Navcam    & 396     & 70      & 70      & 140     \\
\bf Mastcam-Z & 1034    & 70      & 70      & 972     \\
\hline
\end{tabular}
\caption{Number of input RDRs per product type for contextual mesh 0534\_0261222.}
\label{RDRsPerProductType}
\end{table}

\begin{table}
\renewcommand{\arraystretch}{1.3}
\centering
\footnotesize
\begin{tabularx}{\columnwidth}{|r|X|X|X|X|X|}
\hline
              & \bf 0260850 & \bf 0261004 & \bf 0261110 & \bf 0261154 & \bf 0261222 \\
\hline\hline
\bf Hazcam    & 12          & 226         & 0           & 12          & 246         \\
\bf Navcam    & 70          & 70          & 6           & 56          & 387         \\
\bf Mastcam-Z & 342         & 770         & 0           & 96          & 798         \\
\hline
\end{tabularx}
\caption{Number of input RDRs per sitedrive for contextual mesh 0534\_0261222.}
\label{RDRsPerSiteDrive}
\end{table}

The worker downloads that set of RDR input data from the ODS.  It also downloads the orbital DEM (13000x9000, 1 pixel per m, about 450MB) and orthoimage (52000x36000, 4 pixels per m, about 2.4GB) if they aren't yet locally cached on its filesystem, or if they've changed.  These are large files, but the GeoTIFF format supports sparse reads and the files can be locally cached.

The worker downloads all the input data from ODS.  It then ingests relevant metadata from every product into a local database that will be used in all subsequent processing.  This includes the RDR image camera models, coordinate frames, and color models, as well as the mapping from pixel to latitude and longitude in each of the orbital data products.  Pose priors are generated by querying the PLACES service to get the estimated rover latitude, longitude, and elevation at each sitedrive.  The values for contextual mesh 0534\_0261222 are shown in Table~\ref{PositionPriors}.  Corresponding rover frame orientation quaternions from the rover's inertial measurement unit (IMU) are also extracted from the RDR headers.
\begin{table*}
\renewcommand{\arraystretch}{1.3}
\centering
\begin{tabular}{|r|l|l|l|l|l|}
\hline
                             & \bf 0260850 & \bf 0261004 & \bf 0261110 & \bf 0261154 & \bf 0261222 \\
\hline\hline
\bf Sols                     & 474--477    & 477--499    & 501         & 501         & 502--533    \\
\bf Longitude [degrees]      & 77.40609995 & 77.40616768 & 77.40603023 & 77.40595178 & 77.40592495 \\
\bf Latitude [degrees]       & 18.45885590 & 18.45889532 & 18.45872741 & 18.45869497 & 18.45864728 \\
\bf Easting [m]              & 4351976.114 & 4351979.920 & 4351972.189 & 4351967.781 & 4351966.269 \\
\bf Northing [m]             & 1094143.100 & 1094145.436 & 1094135.483 & 1094133.563 & 1094130.735 \\
\bf Elevation [m]            & -2521.591   & -2520.853   & -2522.764   & -2522.828   & -2523.168   \\
\bf $\Delta$X to 0261222 [m] & 12.366      & 14.703      & 4.750       & 2.827       & 0.000       \\
\bf $\Delta$Y to 0261222 [m] & 9.839       & 13.647      & 5.919       & 1.508       & 0.000       \\
\bf $\Delta$Z to 0261222 [m] & 1.577       & 2.315       & 0.404       & 0.340       & 0.000       \\
\bf Distance to 0261222 [m]  & 15.881      & 20.194      & 7.600       & 3.222       & 0.000       \\
\hline
\end{tabular}
\caption{Rover position priors from the PLACES database at each sitedrive for contextual mesh 0534\_0261222, as well as the sol range of corresponding data products.  Rover orientation quaternion priors are also retrieved but not shown here.  \emph{Easting} and \emph{northing} are computed from longitude and latitude: $\text{easting} = (\text{longitude}/360^\circ)C_M\cos{\phi_0}$, $\text{northing} = (\text{latitude}/360^\circ)C_M$ where $C_M=21338891.108$m is taken to be the equatorial circumference of Mars and $\phi_0=18.4663^\circ$ is the standard parallel used for equirectangular projection Mars 2020 mission maps.  Mars 2020 mission surface Cartesian coordinate frames without rotation, i.e. SITE and LOCAL\_LEVEL frames, are typically right handed with $+$Z down, $+$X North, and $+$Y East~\protect\cite{M20PlacesSIS}.}
\label{PositionPriors}
\end{table*}

In practice it typically takes around 30 minutes to download and ingest the inputs for a contextual mesh.  The result of ingestion is a full prior of the camera model intrinsics and extrinsics for every surface observation---both the original left/right images as well as the XYZ/UVW wedge pointclouds from stereo vision processing---in a common coordinate frame that is also registered to the orbital data.

\section{Alignment Algorithms}
In-situ observations within one sitedrive (i.e. acquired from one static rover stopping position) are typically already very well aligned to each other by pose priors from telemetry and the PLACES database because the rover kinematics and cameras are typically well calibrated.  However, the alignment of data across sitedrives may not be as good due to errors in the estimation of the rover's locomotion trajectories, which are typically based on odometry with periodic ground-based manual localization to the orbital image.

Unlike many other photogrammetry and stereo SLAM systems, the Landform contextual mesh does not attempt to adjust the poses of observations within each sitedrive.  However, we do perform several stages of alignment to refine the alignment of sitedrives to each other and to the orbital data.

\subsection{Birds Eye View Feature Matching}
\begin{figure*}
\centering
\includegraphics[width=\textwidth]{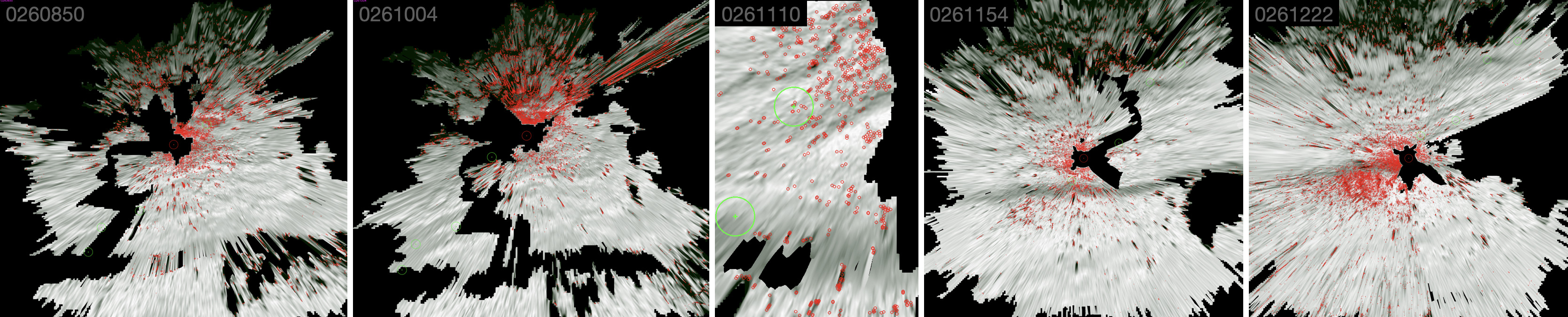}
\caption{FAST features (Features from Accelerated Segment Test)~\protect\cite{Rosten06}, shown as red dots, are detected in each birds-eye view sitedrive image.}
\label{FASTFeatures}
\end{figure*}

The first alignment stage is a sparse feature based algorithm based on \emph{birds eye view} (BEV) renderings of each sitedrive.  The XYZ pointclouds from stereo vision are meshed and rendered in a synthetic top-down view at a fixed resolution of 1cm per pixel, creating one BEV image per sitedrive.  FAST features (Features from Accelerated Segment Test)~\cite{Rosten06} are detected in those images (Figure~\ref{FASTFeatures}) and matched to those for other sitedrives.  The resulting matches are filtered for outliers and then used to refine the alignment between the sitedrives using a form of 3D pose graph optimization.  Though the features are detected in 2D image space, the alignment is still performed in 3D by using the XYZ observation data to map the 2D feature matches to 3D point matches.

To form the per-sitedrive meshes, BEV images, and FAST features we process each sitedrive $d$ in parallel:
\begin{enumerate}
\item For each XYZ wedge pointcloud RDR $r$ in $d$, in parallel batches:
  \begin{enumerate}
  \item Considering $r$ as a 3 band image, re-scale to at most $256 \times 256$ points.  We have found that it's not necessary to use the full resolution of the source data for this algorithm, and this step greatly reduces the subsequent computational requirements.
  \item Build a triangle mesh $m_r$ by organized meshing $r$.
  \end{enumerate}
\item Software rasterize all $m_r$ in $d$, in parallel, in a simulated top-down viewpoint to a \emph{BEV image} $B_d$ for $d$.  Use parallel projection with a scale factor computed to make 1 pixel in the image correspond to 1 square centimeter in the XY plane of the terrain, and false color the image using a colormap proportional to the tilt angle of the terrain.
\item Render a secondary \emph{heightmap image} $H_d$ in the same way, but in this case use a colormap that represents relative elevation on the terrain.
\item The XYZ pointclouds resulting from stereo vision naturally become more sparse with distance from the camera.  This effect transfers to the rendered images $B_d$ and $H_d$: often they will have large peripheral areas of pixels which received no color in the rasterization.  We apply a heuristic algorithm to $B_d$ and $H_d$ (in lock-step) to remove blocks of pixels which have fewer rasterized pixels than a set threshold.
\item Run a FAST feature detector~\cite{Rosten06} on $B_d$.
\end{enumerate}
Coloring the BEV images by terrain slope---not the actual observed terrain appearance---makes the BEV aligner relatively invariant to scene lighting.  Fixing the resolution at 1cm per pixel also adds a degree of scale invariance by limiting the minimum scale of detected features: small details on the terrain are ignored but larger and more significant features, such as rocks and ridges, remain.

\begin{figure}
\centering
\includegraphics[width=\columnwidth]{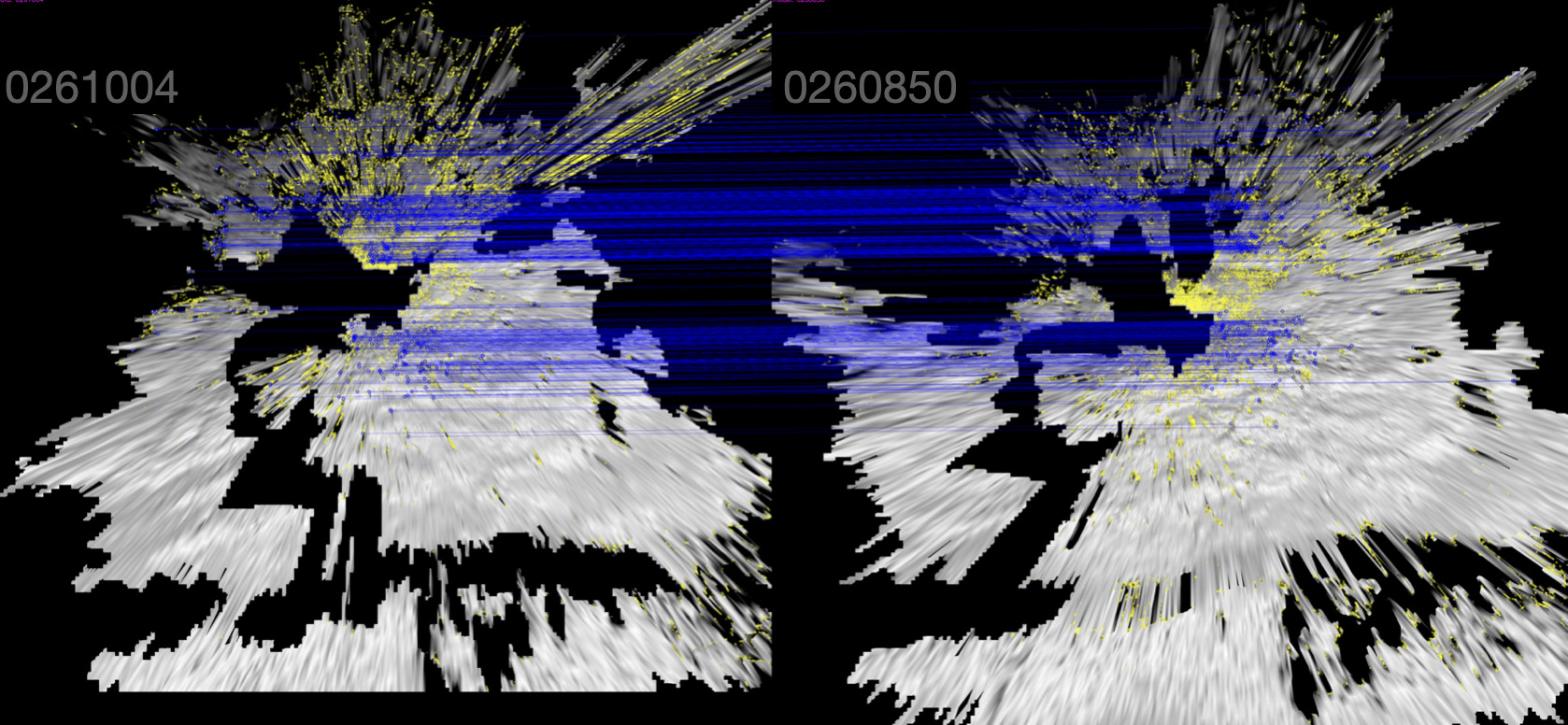}
\caption{Feature matches between sitedrives 02610045 and 0260850 that survived outlier rejection.}
\label{FigMatches}
\end{figure}

The next algorithm finds pairs of matching features between BEV images and filters the matches to reject outliers (Figure~\ref{FigMatches}).  For each pair $(i, j)$ of sitedrives (in parallel):
\begin{enumerate}
\item Use the relative $(\Delta x, \Delta y)$ priors from PLACES (Table~\ref{PositionPriors}), converted from meters to pixels using the fixed scale factor $1\text{pixel}=1\text{cm}$, to initially position $B_i$ relative to $B_j$.
\item Compute a scale-invariant feature transform (SIFT) descriptor~\cite{Lowe99} for each FAST feature in $B_i$ and $B_j$.
\item For each feature $a\in B_i$, collect all features $b\in B_j$ where
  \begin{itemize}
  \item $\|\text{center}(a)-\text{center}(b)\| < 100$ pixels \emph{and}
  \item the distance between $a$ and $b$ in descriptor space is less than a threshold.
  \end{itemize}
\item Apply random sample consensus (RANSAC)~\cite{Fischler81} to the feature pairs $(a, b)$.  Repeat up to $\text{min}\left(5000000, \binom n2\right)$ times or until the best residual $R$ is less than 2cm:
  \begin{enumerate}
  \item Pick two seed feature matches $(a_1, b_1)$, $(a_2, b_2)$.  We use either a random shuffle of all $\binom n2$ feature combinations $(a_1, a_2)$ or random sampling with duplicate rejection if the shuffle would be too large.
  \item Use the seed matches to compute a 2D rigid transform $T_s$ that moves $B_j$ relative to $B_i$ so that $b_1$ aligns to $a_1$ and $b_2$ aligns to $a_2$ as closely as possible.
  \item Temporarily move $B_j$ by $T_s$ and find all feature matches $M$ for which $\|\text{center}(a)-\text{center}(b)\| < 5$ pixels.
  \item If $|M|\geq 25$:
    \begin{enumerate}
    \item Re-estimate 2D rigid transform $T_M$ using all matches in $M$ by orthogonal Procrustes~\cite{Gower04}.
    \item Temporarily move $B_j$ by $T_M$ and compute the residual $$r_M=\sum_{(a,b)\in M}\|\text{center}(a)-\text{center}(b)\|\left(\frac{1\text{cm}}{1\text{pixel}}\right)$$.
    \item Save $M$ and $T_M$ if $r_M$ is the lowest residual found so far.
    \end{enumerate}
  \end{enumerate}
\item Transform $B_j$ by the best transform $T_M$ found by RANSAC, and keep only the corresponding matches $M$.
\item Convert the 2D features used in the selected matches to 3D points.  The $x$ and $y$ coordinates are simply converted from pixels to meters by the fixed scale factor of $1\text{cm}=1\text{pixel}$.  The elevation map images $H_i$ and $H_j$ are used to compute the $z$ coordinates along with the relative $\Delta z$ priors from PLACES (Table~\ref{PositionPriors}).
\item Discard outlier matches separated by a spatial distance greater than 5 absolute deviations from the median~\cite{Leys13}.
\item Give up on aligning $i, j$ if less than 25 matches remain.
\end{enumerate}

\begin{figure}
\begin{subfigure}{0.495\columnwidth}
    \includegraphics[width=\columnwidth]{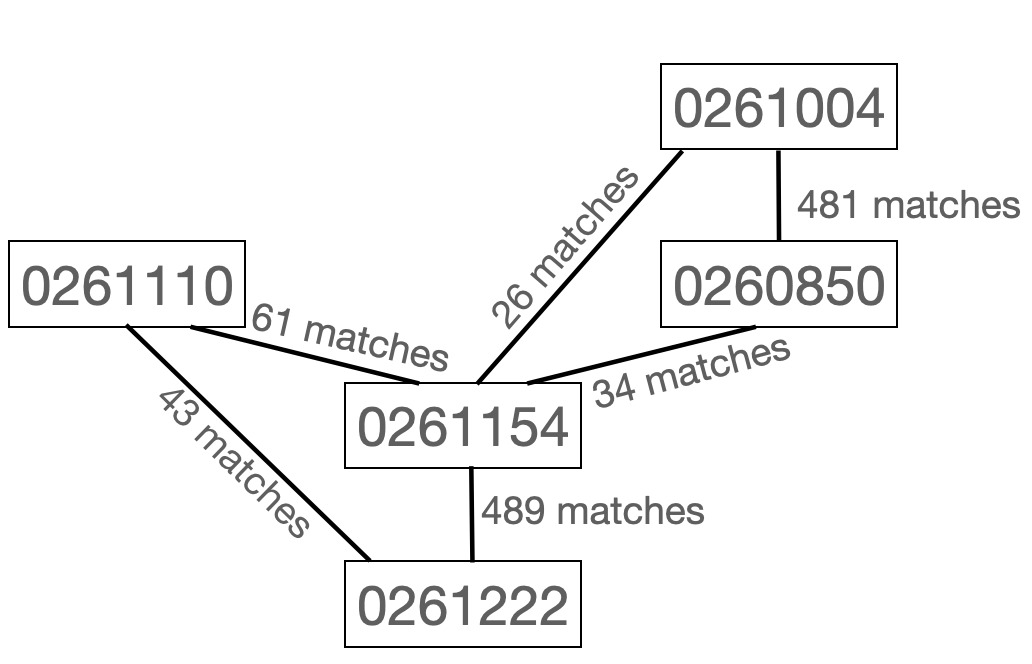}
\end{subfigure}
\hfill
\begin{subfigure}{0.495\columnwidth}
    \includegraphics[width=\columnwidth]{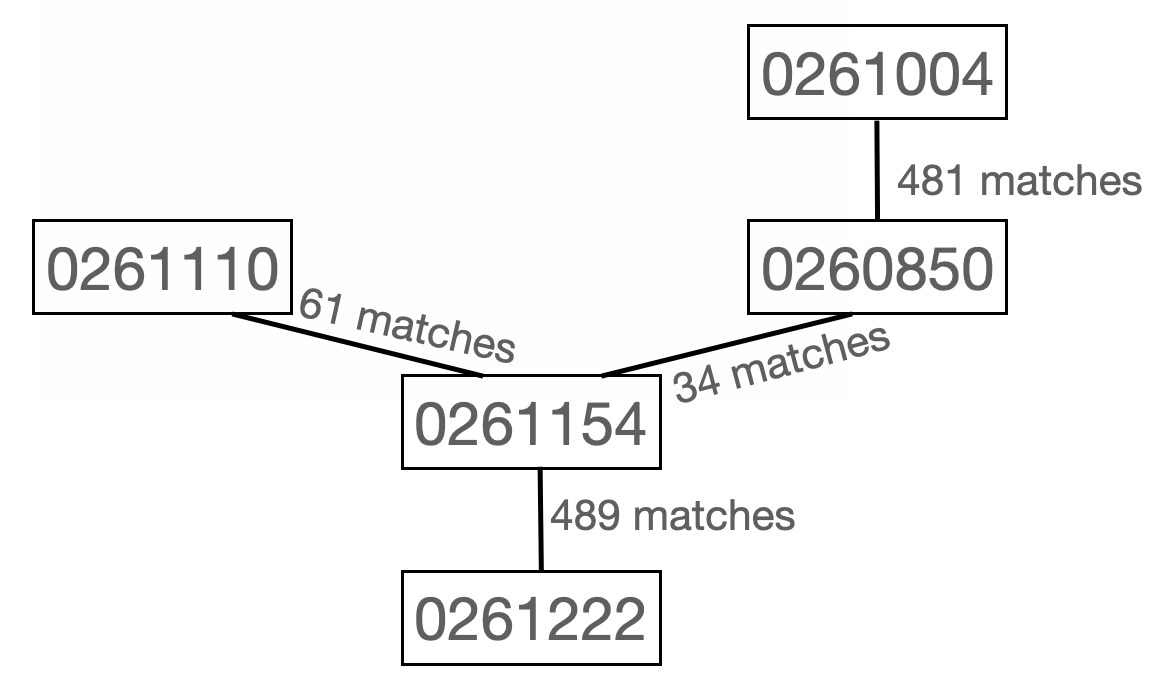}
\end{subfigure}
\caption{BEV aligner pose graph and corresponding spanning tree for contextual mesh 0534\_0261222.}
\label{FigBEVGraph}
\end{figure}

\begin{figure}
\begin{subfigure}{\columnwidth}
\includegraphics[width=\columnwidth]{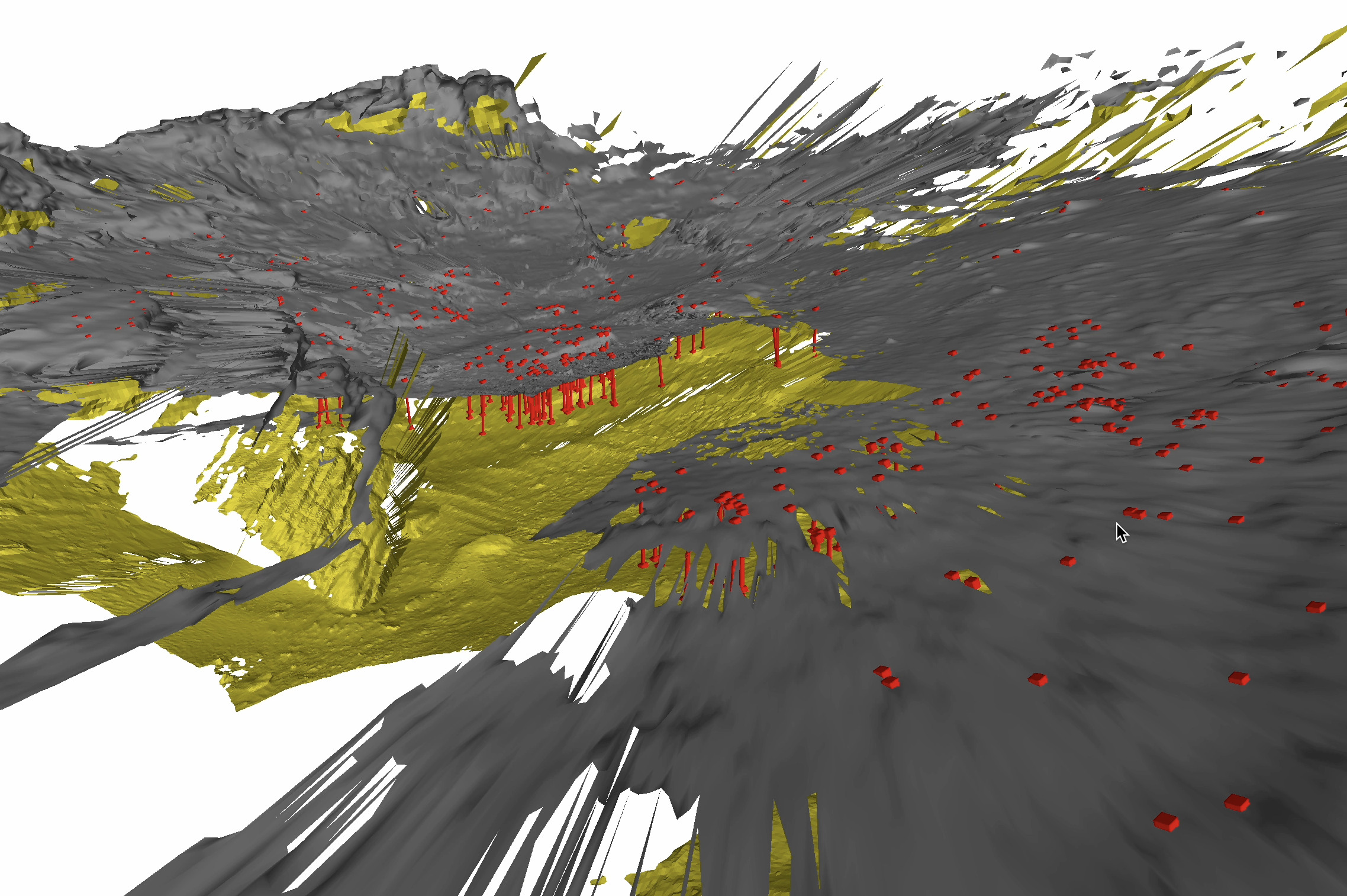}
\end{subfigure}
\begin{subfigure}{\columnwidth}
\includegraphics[width=\columnwidth]{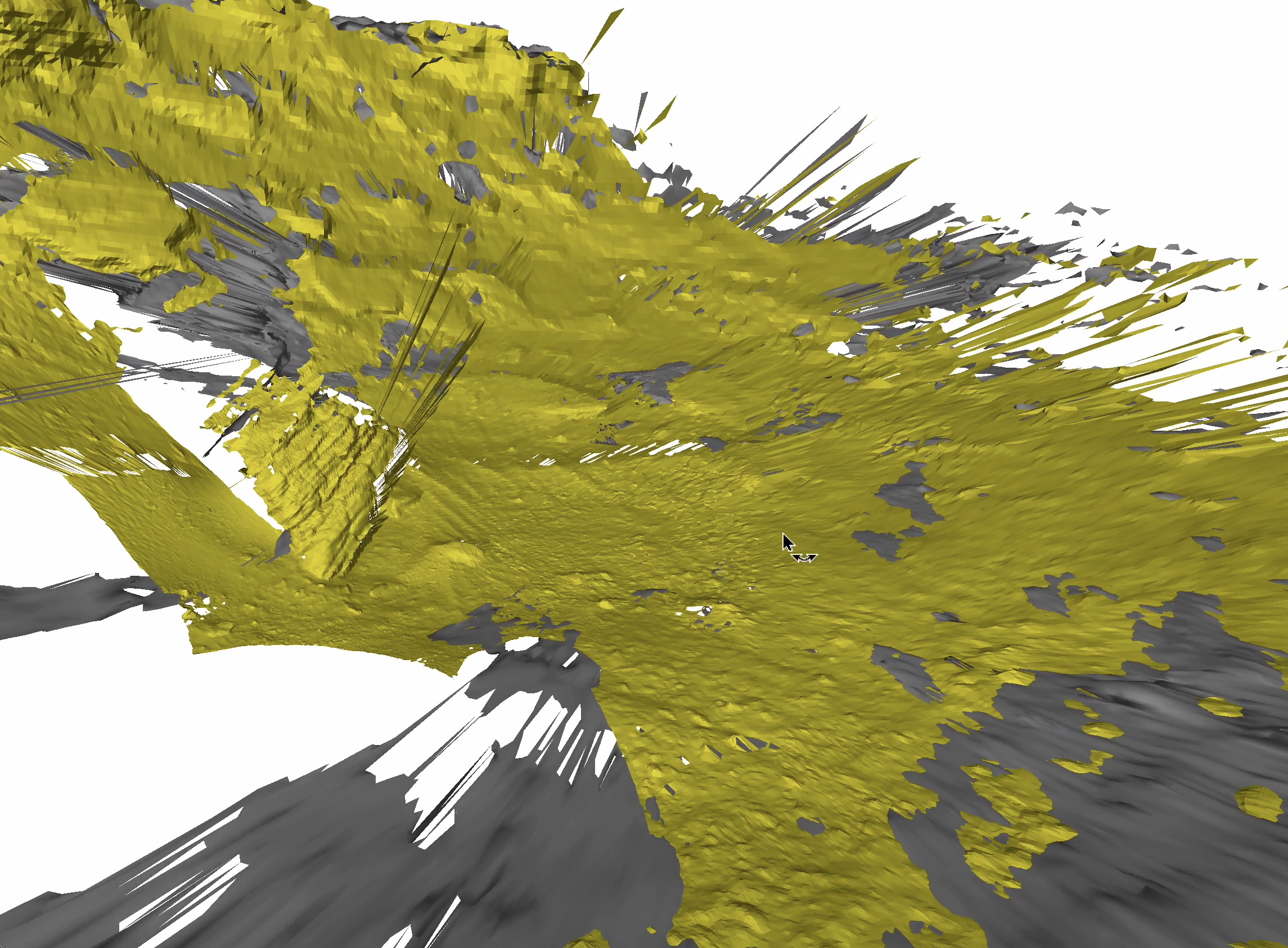}
\end{subfigure}
\caption{Feature matches for sitedrives 02610045 and 0260850 converted to 3D points (top) and result of BEV aligner after minimizing the corresponding residual (bottom).  In this case the residual was reduced from 0.49m to 0.037m.}
\label{FigBEV}
\end{figure}

The collections of feature matches surviving outlier rejection, if any, between each pair of sitedrives can be considered edges in a graph of the sitedrives, as shown in Figure~\ref{FigBEVGraph} (left).  This can also be considered a \emph{pose graph} where each sitedrive is a 6 degree of freedom rigid body in 3D, and the edges are point correspondences between those bodies.
Existing algorithms can adjust the poses of all nodes in such a graph to optimize the total residual of the point correspondences~\cite{Kummerle11}.  We have not yet integrated such an algorithm.  Instead we currently use a non-optimal approach that we have found is still effective in practice.  We simply find a spanning tree of the pose graph (Figure~\ref{FigBEVGraph}, right).  The remaining tree is then optimized by applying orthogonal Procrustes independently to each pair of sitedrives connected by a surviving edge.  Figure~\ref{FigBEV} shows an example.  It would be better to optimize the entire original graph---the spanning tree approach is akin to ignoring loop closures in traditional SLAM---but that's a loose end in our implementation that remains future work.

The BEV aligner typically takes about 10-30 minutes.

\subsection{Heightmap Alignment with ICP}
Typically the BEV aligner makes a significant improvement in the relative alignment of sitedrives.  However, in some cases there may be too few successful feature matches.  Also, we haven’t yet aligned the sitedrives (i.e. the in-situ data from rover-based stereo vision) to the orbital DEM.

\begin{figure}
\centering
\includegraphics[width=\columnwidth]{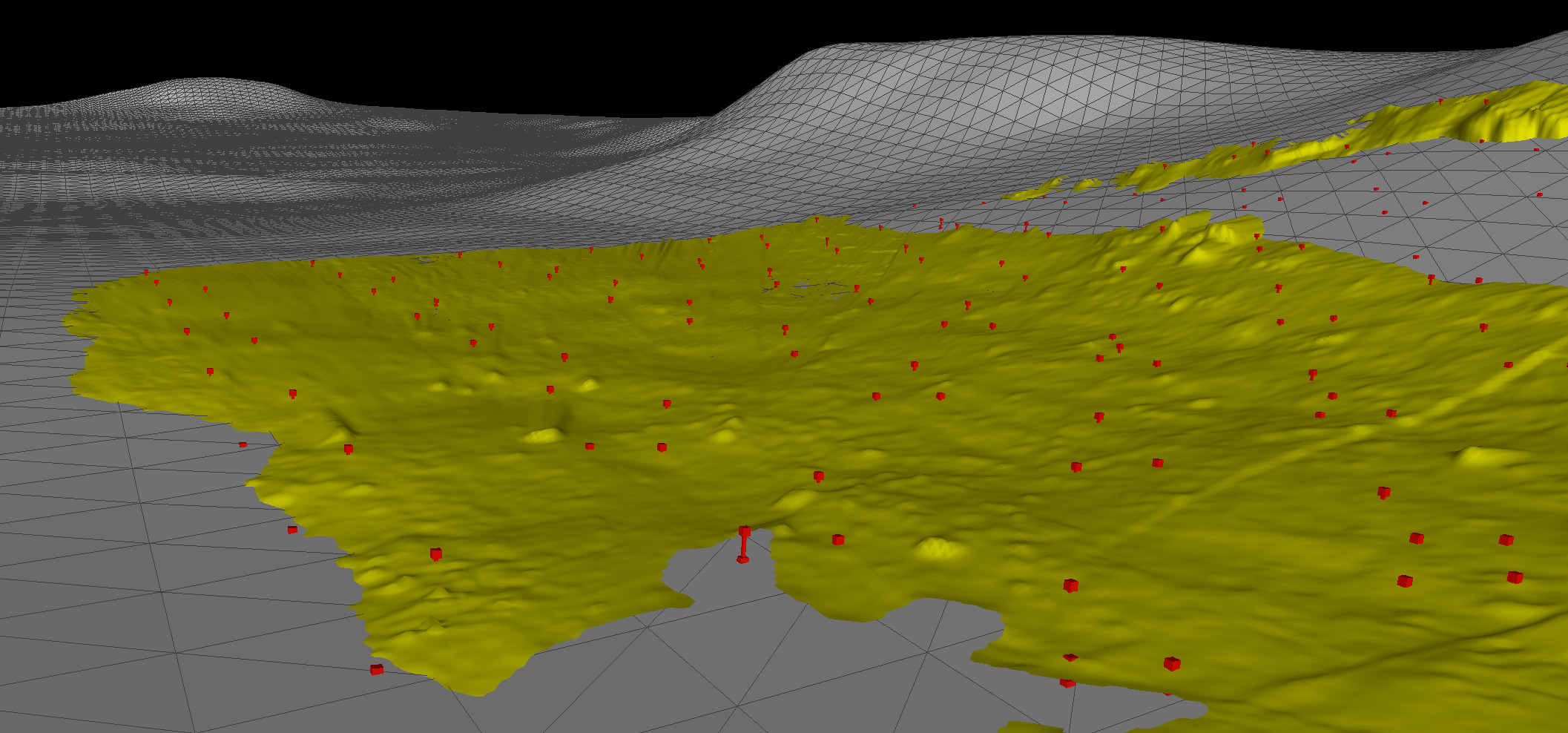}
\caption{Iterative closest point (ICP) alignment.  This image shows the in-situ sitedrive data being aligned to the orbital DEM.  The ICP aligner also refines the alignment of sitedrives relative to each other.}
\label{FigICP}
\end{figure}

To address these issues, we now apply a second alignment algorithm based on iterative closest point (ICP)~\cite{Arun87}.  This algorithm uses same sitedrive meshes that were built for the BEV aligner, but uses a grid of vertically corresponding point pairs between sitedrives and also between sitedrives and the orbital DEM, as shown in Figure~\ref{FigICP}.

The heightmap alignment algorithm is:
\begin{enumerate}
\item Initialize aligned group $G$ to contain the central sitedrive for the contextual mesh.
\item Sort the other sitedrives $O$ by decreasing area of their sitedrive meshes.
\item For each sitedrive $s\in O$:
  \begin{enumerate}
    \item Use the aligned pose of $s$ from BEV alignment, if any, as the prior, otherwise use the PLACES prior.
    \item Apply ICP to attempt align $s$ to all the sitedrives in $G$.
    \item If successful, move $s$ from $O$ to $G$.
  \end{enumerate}
\item Align the orbital DEM to $G$ starting with the PLACES prior for the central sitedrive.
\item Add the orbital DEM to $G$.
\item If any sitedrives are left in $O$, align them to $G$---this will typically succeed now that $G$ contains the orbital DEM.
\end{enumerate}

The specific variant of ICP we use is approximate and formulated to align a 3D mesh $A$ to a group of other 3D meshes $B$.  It repeats the following loop until terminated:
\begin{enumerate}
\item Sample a sparse set $S$ of points on meshes in $B$ in a regular spatial grid.  Sample density is automatically chosen to distribute about 1000 samples across all the meshes in $B$.
\item For each point $p_j\in S$ approximate a closest point $q_j$ on $A$ along the vertical line $l_j$ containing $p_j$.  If there is no such point $q_j$ (i.e. $l_j$ did not intersect $A$) then remove $p_j$ from $S$.
\item Remove outliers $p_j\in S$ for which the corresponding $q_j$ is more than 20 absolute deviations away from the mean of $\|p_j-q_j\|$ for all point pairs $(p_j, q_j)$.
\item Define the current residual $r_i$ as the sum of the distances $\|p_j-q_j\|$ for all point pairs $(p_j, q_j)$.
\item Terminate if
  \begin{itemize}
  \item the current residual is below a threshold \emph{or}
  \item a maximum number of iterations has been reached \emph{or}
  \item $r_i > r_{i-1}$ is greater than the last iteration (diverging).
  \end{itemize}
\item Apply orthogonal Procrustes~\cite{Gower04} to compute a rigid transform $T_i$ that moves all points $p$ towards their paired points $p$ to minimize the residual $r_i$.
\item Apply $T_i$ to move $A$ closer to $B$.
\end{enumerate}

Heightmap alignment for a contextual mesh typically takes a few minutes or less.

\section{Meshing}
The next stage after aligning all sitedrives to each other and to the orbital DEM is to create a triangle mesh that represents the terrain geometry.  This mesh will cover the full 1km square extent of the contextual mesh and will be the basis for creating the terrain tileset.

We begin by combining all the points from the XYZ and UVW RDRs, creating a single large combined pointcloud with points and normals.  For contextual mesh 0534\_0261222 this combined point cloud has about 30 million points and covers an area of about 110$\times$70m on the terrain.  The density of points varies greatly---stereo vision produces high density samples close to the observing camera but increasingly lower density with distance.  Also, some areas will be covered by multiple overlapping observations, while other areas may have only one or even no observations.

We compute the convex hull of the projection of the point cloud in the XY plane, giving a convex polygon boundary $B$.  We then augment the point cloud by adding in additional samples from the orbital DEM within that boundary, oversampling it at 8 points per meter.  This adds about 500,000 additional points for 0534\_0261222, which help to fill holes, and which also help guide subsequent mesh reconstruction towards the geometry of the orbital DEM, particularly near the edges of the pointcloud.

We next apply a voxel based algorithm that thins out the higher density regions of the pointcloud.  This algorithm prioritizes keeping points that were closer to the observing camera, as they would typically have higher certainty.  For 0534\_0261222 this reduces the size of the combined pointcloud to about 6 million points.

\begin{figure}
\centering
\includegraphics[width=\columnwidth]{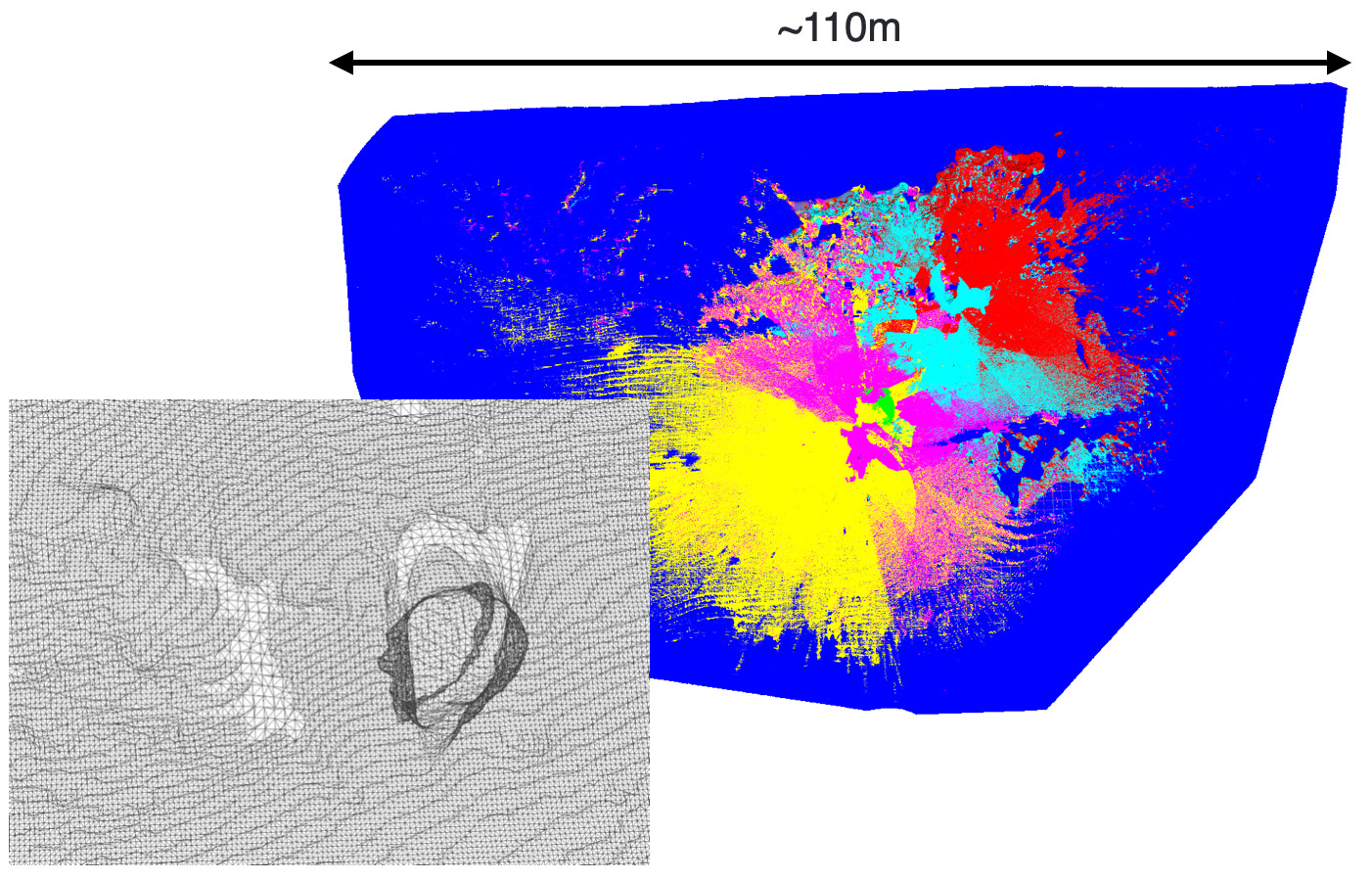}
\caption{Poisson reconstruction~\protect\cite{KazhdanPoisson} is used to build a triangle mesh in the detailed central area covered by in-situ rover observations.  The result is a full 3D mesh, so unlike the orbital DEM, it may represent geometry with overhangs. Poisson also produces an adaptive density triangle mesh, compared to the fixed density organized meshing we use for the outer orbital area.  And it can handle the irregular sampling, noise, and outliers of the combined surface data pointcloud.}
\label{SurfaceMesh}
\end{figure}

We then apply Poisson reconstruction~\cite{KazhdanPoisson} to generate a triangle mesh which approximates the surface.  The Poisson algorithm is capable of handling relatively large pointclouds with noise and outliers.  It generates an adaptive triangle mesh, shown in Figure~\ref{SurfaceMesh}, that represents the local terrain features.  The raw output from Poisson reconstruction often includes ``wings'' that extend beyond the original data---we remove those by clipping to the convex hull boundary $B$.  For 0534\_0261222 this results in a mesh with about 6 million triangles.

\begin{figure}
\centering
\includegraphics[width=\columnwidth]{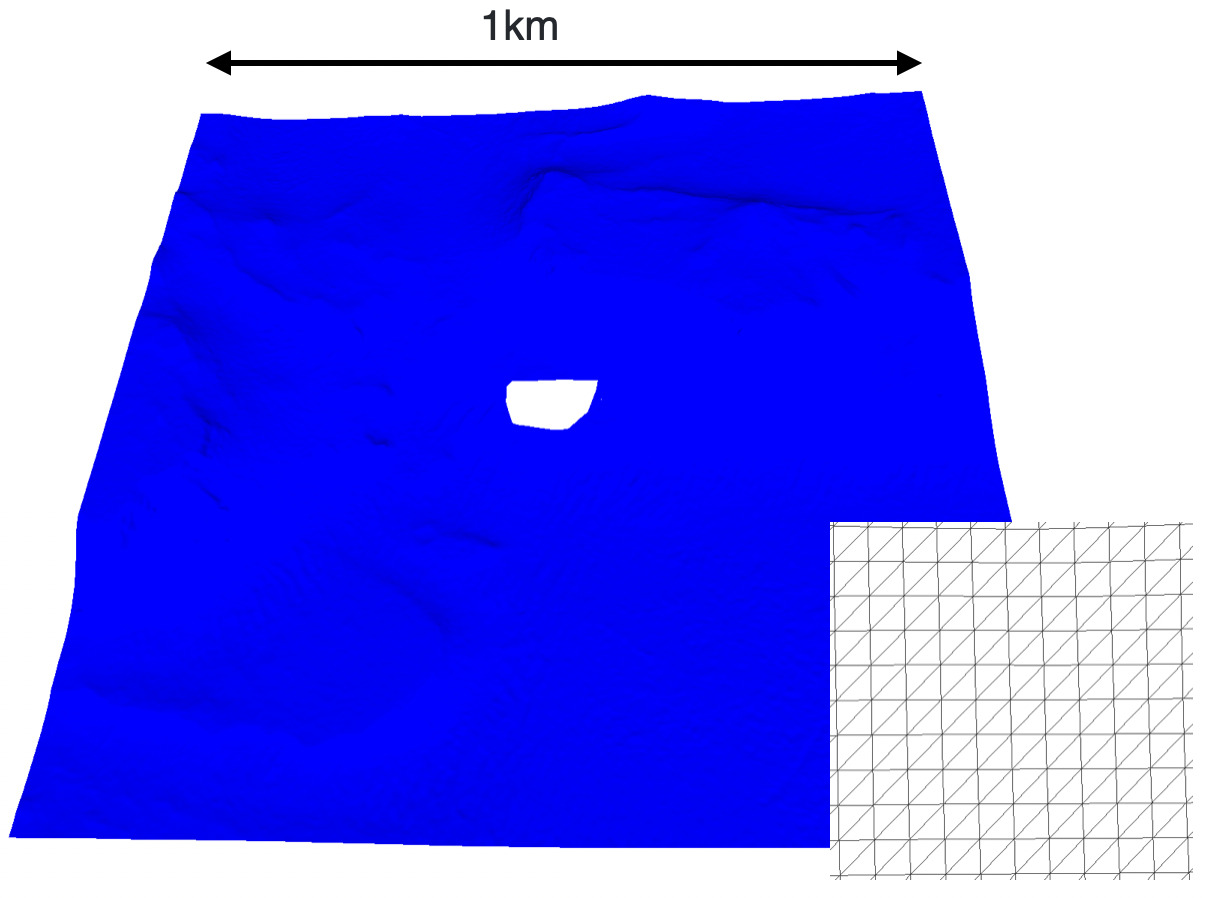}
\caption{The remainder of the full mesh outside the central detail area (Figure~\protect\ref{SurfaceMesh}) is built by sampling the orbital DEM.  By construction the resulting point cloud is only 2.5 dimensional (no overhangs) and is also organized as a regular grid, making it easy to build a triangle mesh for this region.}
\label{OrbitalMesh}
\end{figure}

In order to extend the mesh out to 1km we add more data from the orbital DEM in two parts.  First, in a central square region that is about 3m larger on all sides than the clipped Poisson mesh, we create a \emph{blend mesh} that is oversampled from the orbital DEM at 8 points per meter.  This mesh is easy to create with a simple organized meshing algorithm: the DEM is a 2.5 dimensional heightmap with simple 2-manifold topology, and we take samples of it on a regular grid.  Though the DEM itself only has a resolution of 1m per sample, we intentionally oversample by a factor of 8 to create finer triangles in this region.  We discard those which overlap the convex hull boundary $B$, creating an approximate cut-out for the Poisson mesh, as shown in Figure~\ref{OrbitalMesh}.  We then add a second section of mesh created by sampling the orbital DEM at its native resolution of 1 point per meter, starting from the boundary of the blend mesh and continuing out to the full 1km extent of the contextual mesh terrain.  For 0534\_0261222 the outer orbital mesh has about 2.1 million triangles and the inner blend mesh has about 1.3 million triangles.

\begin{figure*}
\centering
\begin{subfigure}{0.49\textwidth}
    \includegraphics[width=\textwidth]{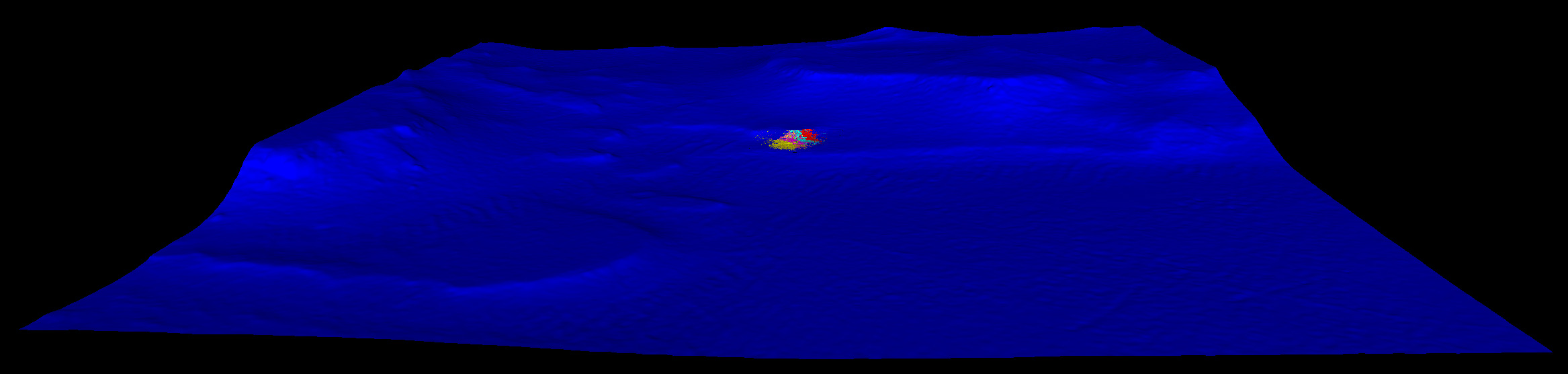}
\end{subfigure}
\hfill
\begin{subfigure}{0.307\textwidth}
    \includegraphics[width=\textwidth]{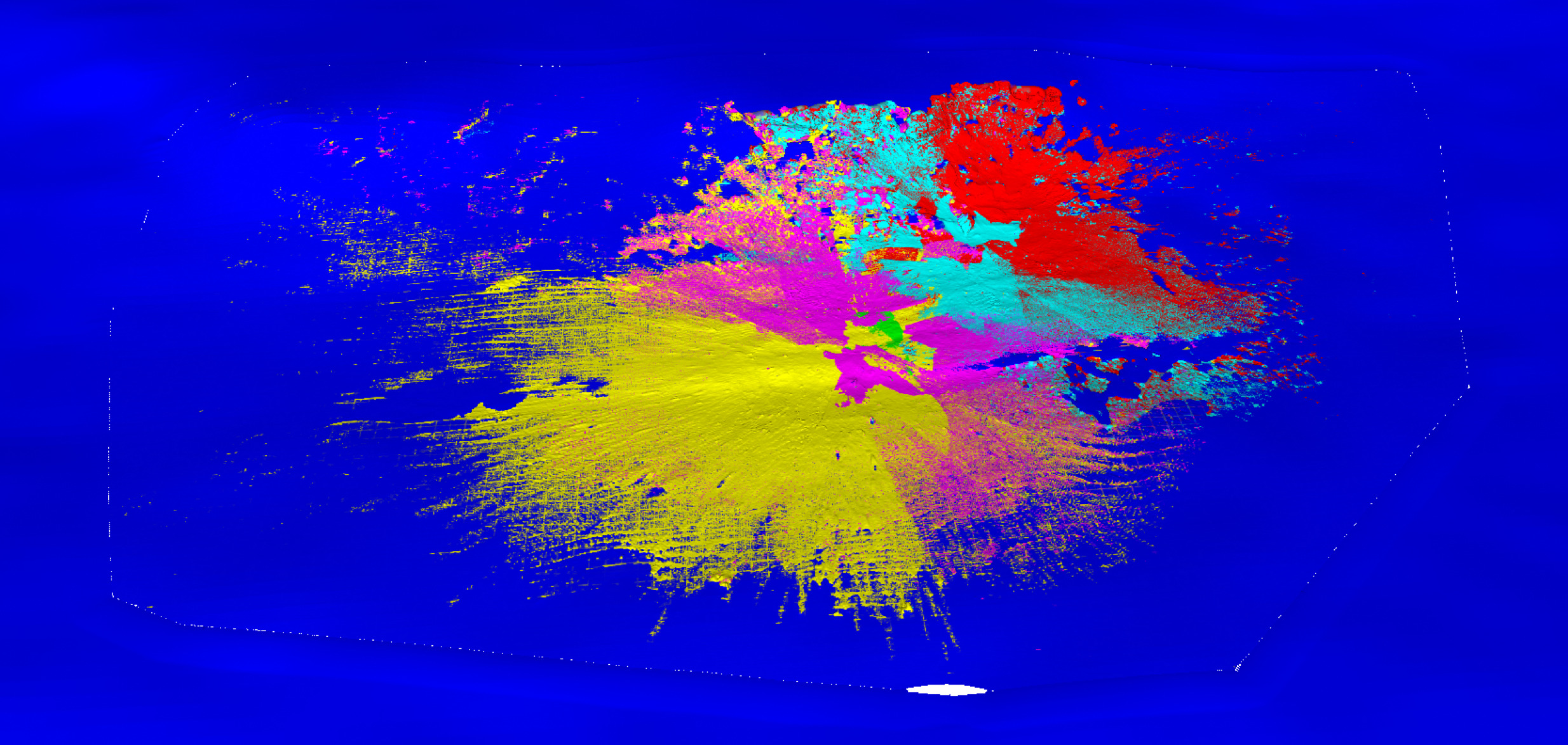}
\end{subfigure}
\hfill
\begin{subfigure}{0.193\textwidth}
    \includegraphics[width=\textwidth]{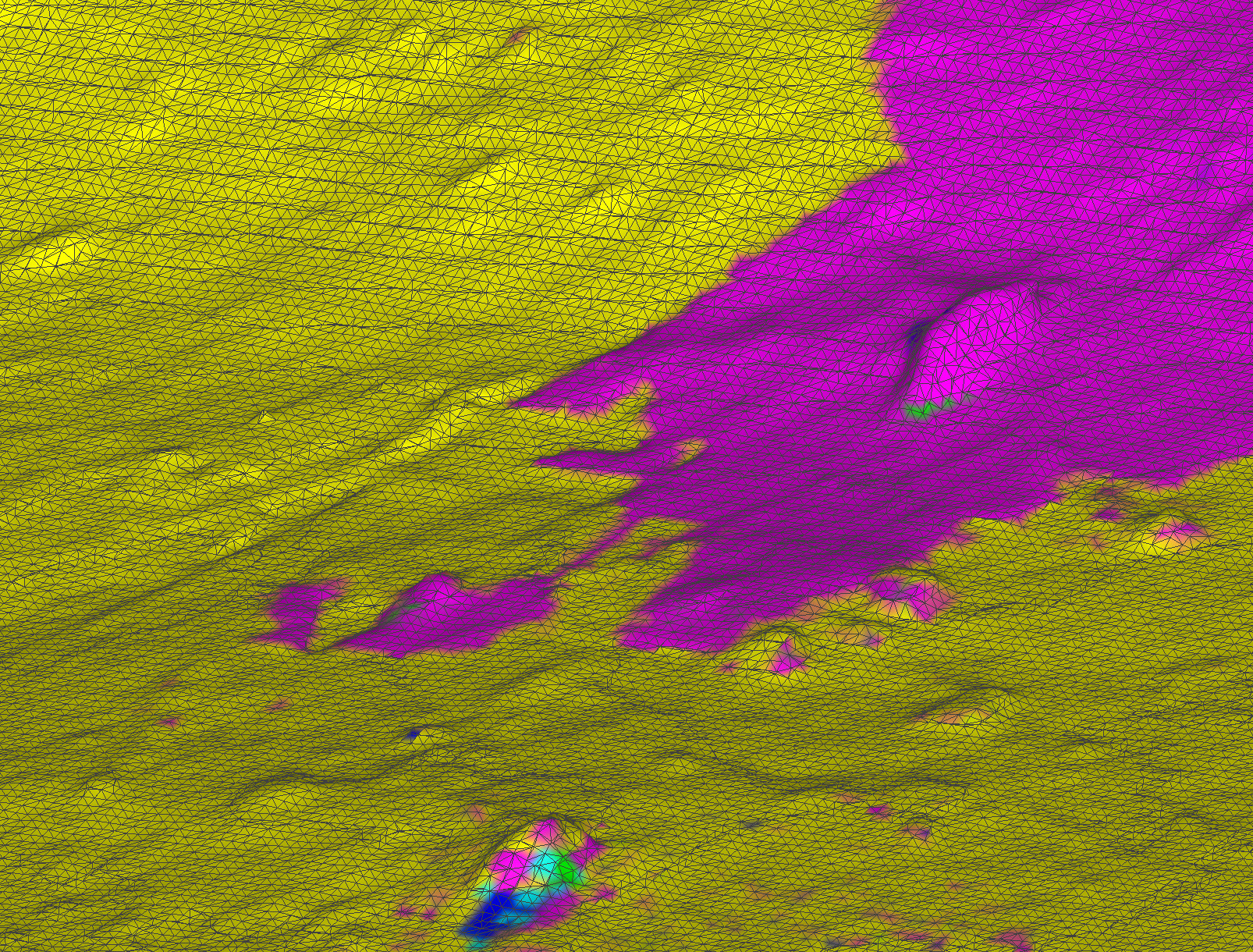}
\end{subfigure}
\caption{Full reconstructed mesh for 0534\_0261222.  Colors indicate the source of the data---blue areas are orbital, other colors  correspond to the 5 sitedrives that were included in this contextual mesh.  Increasing zoom levels from left to right, starting at the full 1km extent.  A topological artifact due to our simple sewing algorithm is visible near the bottom of the middle image.}
\label{FullMesh}
\end{figure*}

The final step in creating the full mesh is to close the gaps between the blend mesh and the Poisson mesh.  We apply an algorithm that (i) adjusts the heights of blend mesh vertices to a weighted average of their neighbors, including those in the Poisson mesh; and (ii) adds \emph{sewing} triangles between nearby boundary vertices in the Poisson and blend meshes.  Robustly sewing across gaps in triangle meshes is nontrivial, and we don't claim that our implementation is highly robust (a defect is visible, for example, in Figure~\ref{FullMesh}, center).  This is an area for future improvement, though we also have observed that in practice the defects are rarely noticeable and don't usually impact the use of the contextual mesh for its intended purpose of situational awareness.
\begin{figure}
\centering
\includegraphics[width=\columnwidth]{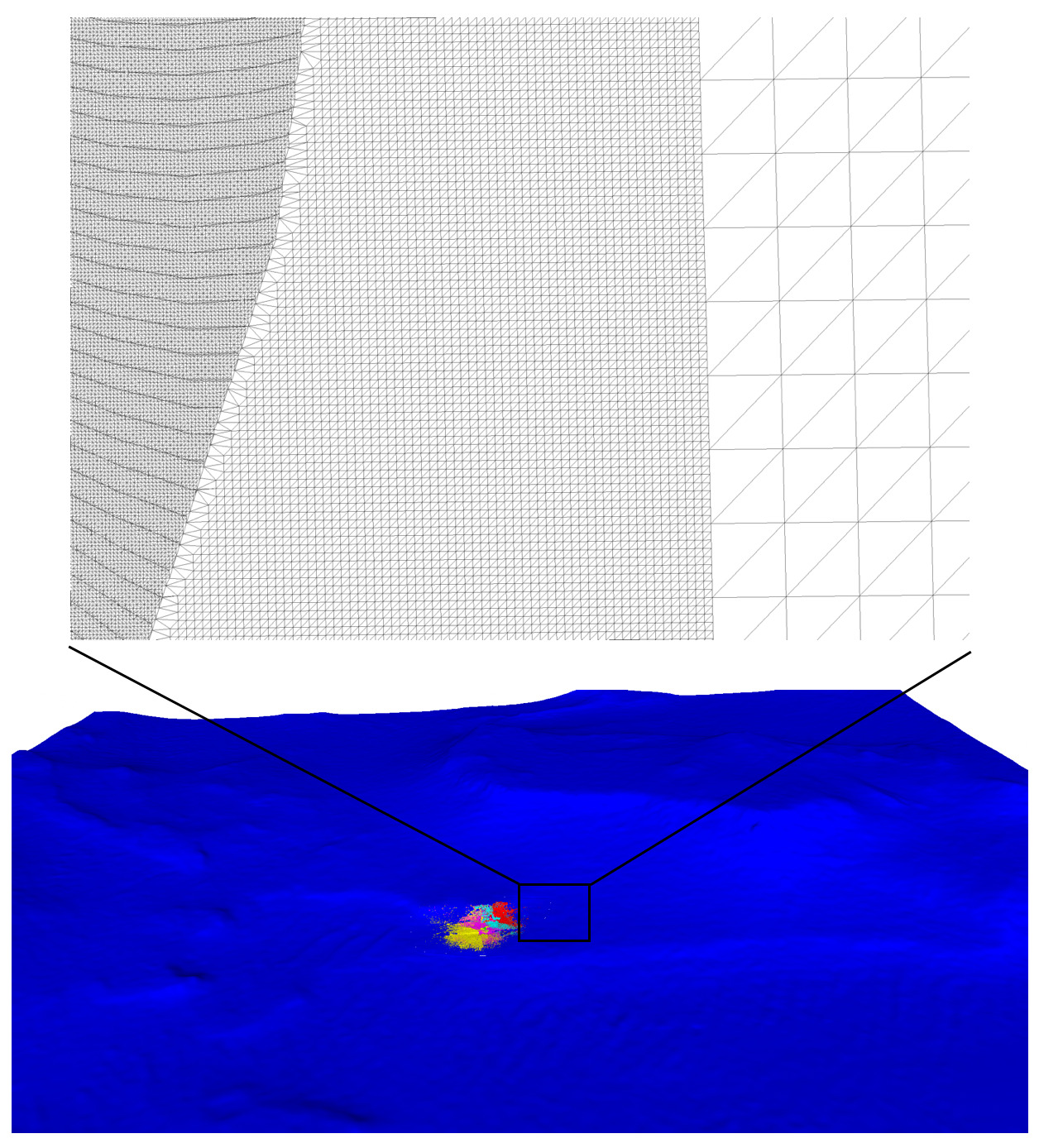}
\caption{Blending and sewing the surface mesh to the orbital mesh.  The wireframe inset shows a section of the surface mesh from Poisson reconstruction (finest triangles at left), the orbital blend mesh (organized medium size triangles in the middle), and the outer orbital mesh (coarse organized triangles at right).  The strip of sewing triangles is between the Poisson mesh and the blend mesh.}
\label{SurfaceOrbitalSewing}
\end{figure}

Our approach of applying Poisson reconstruction where in-situ data is available with organized meshing for the rest of the mesh was developed to balance quality and efficiency.  It would also be possible, for example, to apply Poisson reconstruction to the entire extent.  That might result in fewer artifacts at the periphery of the in-situ data, but would typically require more computational resources. 

Building the full mesh from the aligned RDR pointclouds and the orbital DEM typically takes about 10-30 minutes in practice.  Figure~\ref{FullMesh} shows the results for 0534\_0261222.

\subsection{Mesh Subdivision (Leaf Tiles)}
The full reconstructed mesh is typically on the order of around 10 million triangles, and would often be too large for use in a web-based viewer like ASTTRO.  Clients typically connect with limited bandwidth via the mission virtual private network (VPN).  They are often the laptop machines of mission scientists, and have limited hardware resources.  Another significant limitation is that modern web browsers typically bound the total amount of memory that can be used in each tab.  Also, though $\sim$10M triangles might on its own be tractable in some cases, the corresponding texture image for a monolithic mesh would often have to be extremely large to capture the fine details available in areas with high resolution in-situ rover observations but also have sufficient size to cover the full 1km square extent of the contextual mesh.

To address these constraints we don't simply transfer the full $\sim$10M triangle reconstructed mesh to clients.  Instead, we break it up into a set of typically thousands of tiles in the open standard 3DTiles format~\cite{Cesium3DTiles}.  For example, the tileset for 0534\_0261222 has about 4,000 tiles.  Our approach to do this is integrated with the remaining stages of computation, in particular so that we never have to construct any single super high resolution texture image for the entire mesh.

\begin{figure*}
\centering
\begin{subfigure}{0.245\textwidth}
    \includegraphics[width=\textwidth]{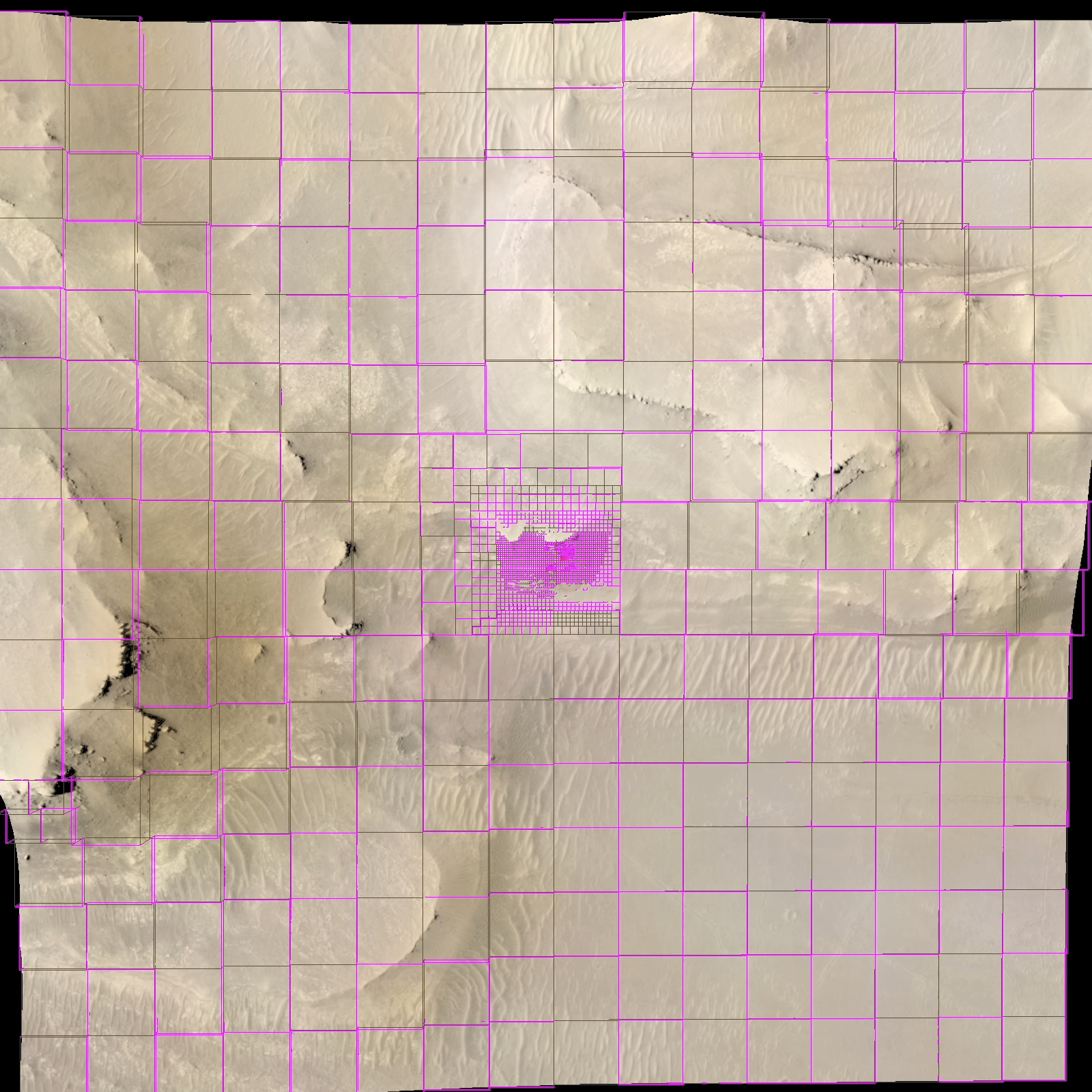}
\end{subfigure}
\hfill
\begin{subfigure}{0.245\textwidth}
    \includegraphics[width=\textwidth]{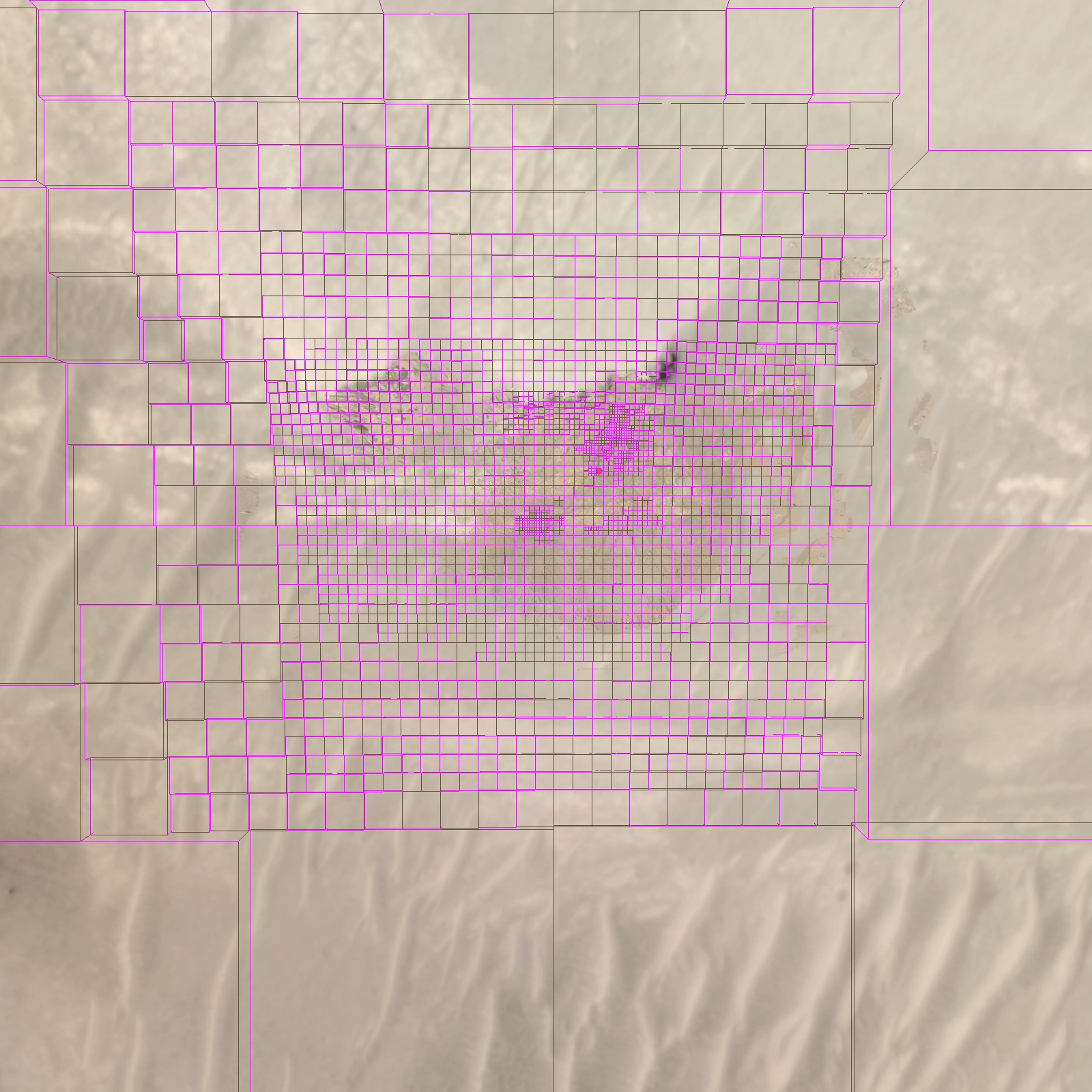}
\end{subfigure}
\hfill
\begin{subfigure}{0.245\textwidth}
    \includegraphics[width=\textwidth]{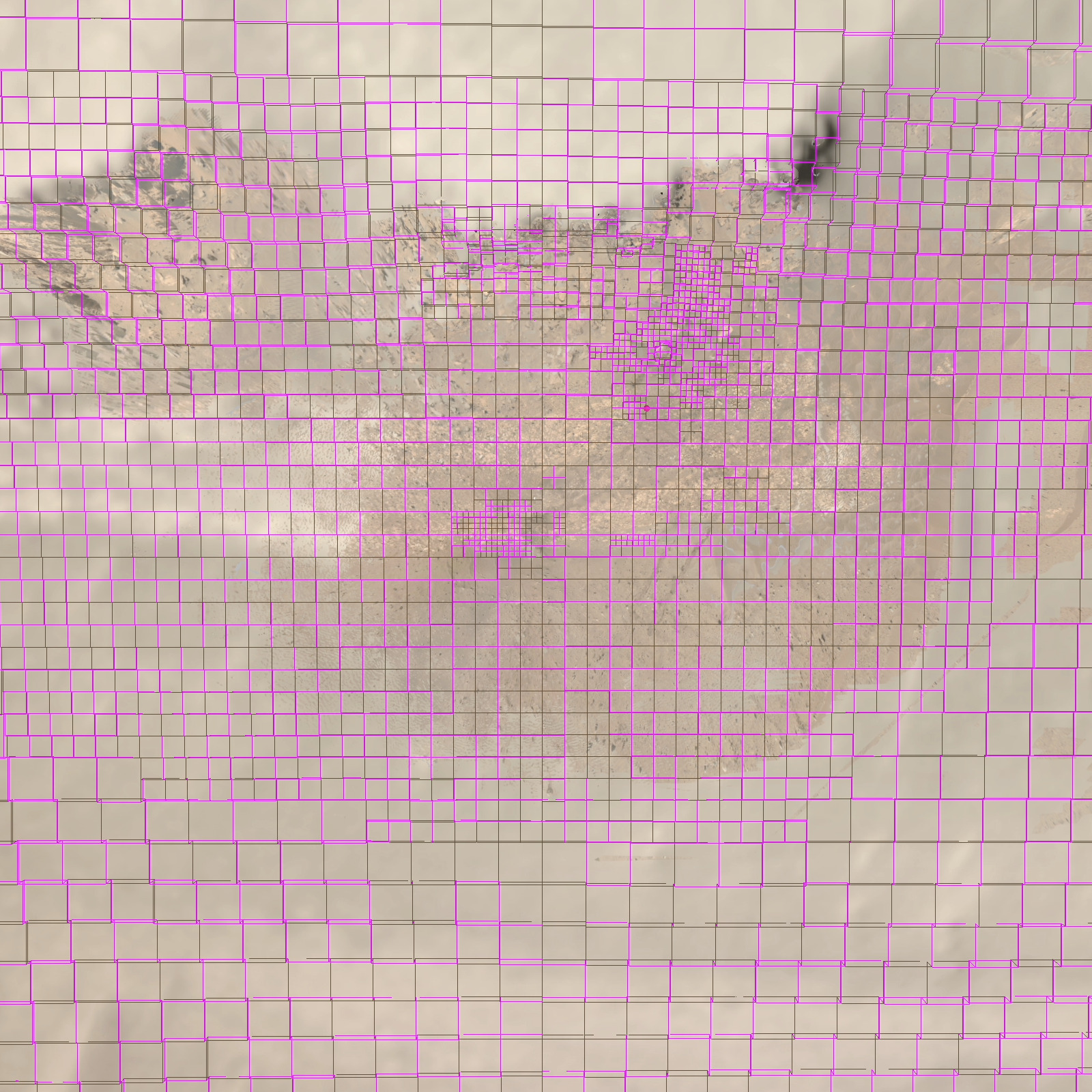}
\end{subfigure}
\hfill
\begin{subfigure}{0.245\textwidth}
    \includegraphics[width=\textwidth]{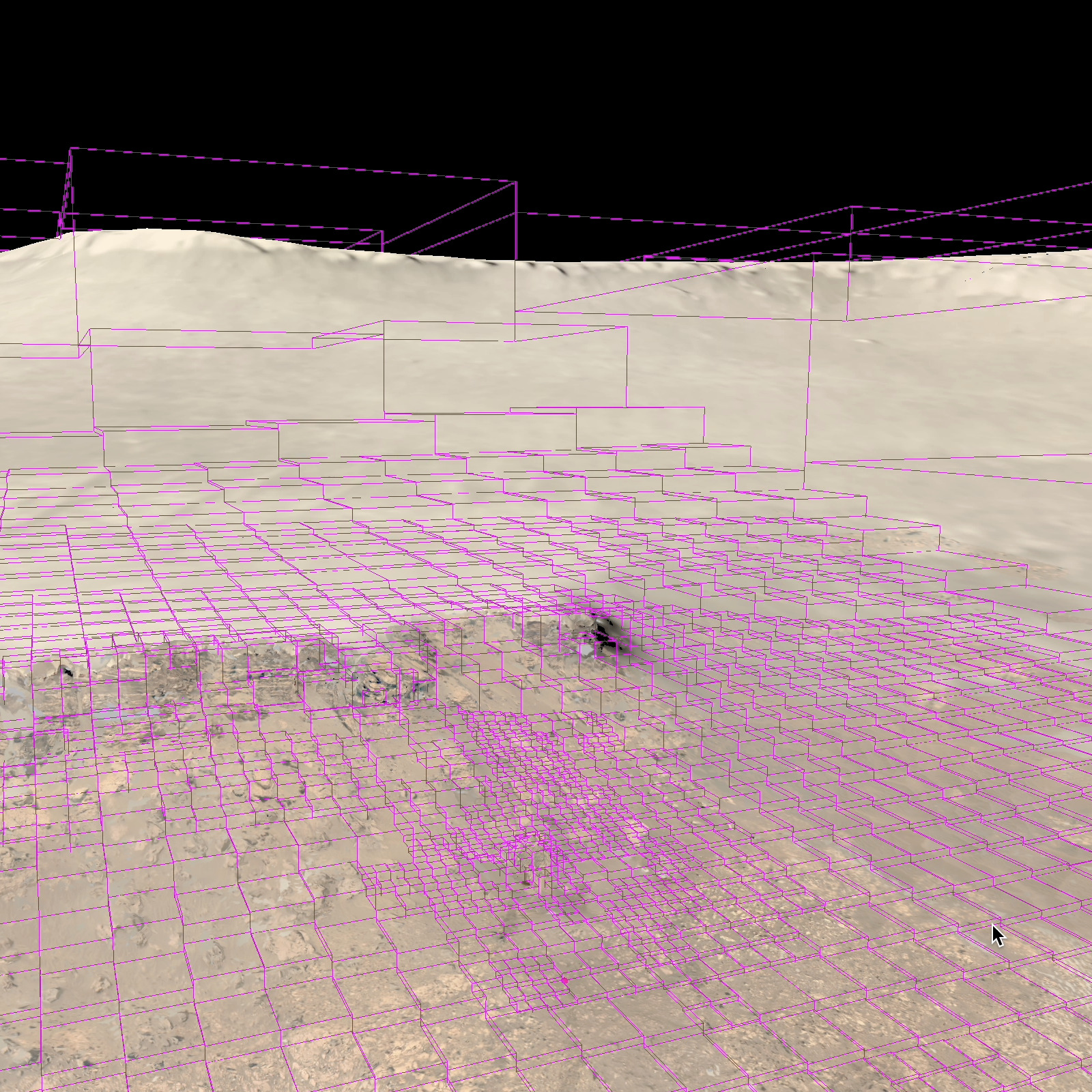}
\end{subfigure}
\caption{Adaptive resolution leaf tiles, formed by subdividing a quadtree to different depths depending on the available resolution of source data in each area.  The left three images show progressively smaller areas in a top-down view, starting from the full 1km square extent of the contextual tileset 0534\_0261222 at left.  The rightmost image shows an oblique view.}
\label{LeafTiles}
\end{figure*}

\begin{figure*}
\centering
\begin{subfigure}{0.34\textwidth}
    \includegraphics[width=\textwidth]{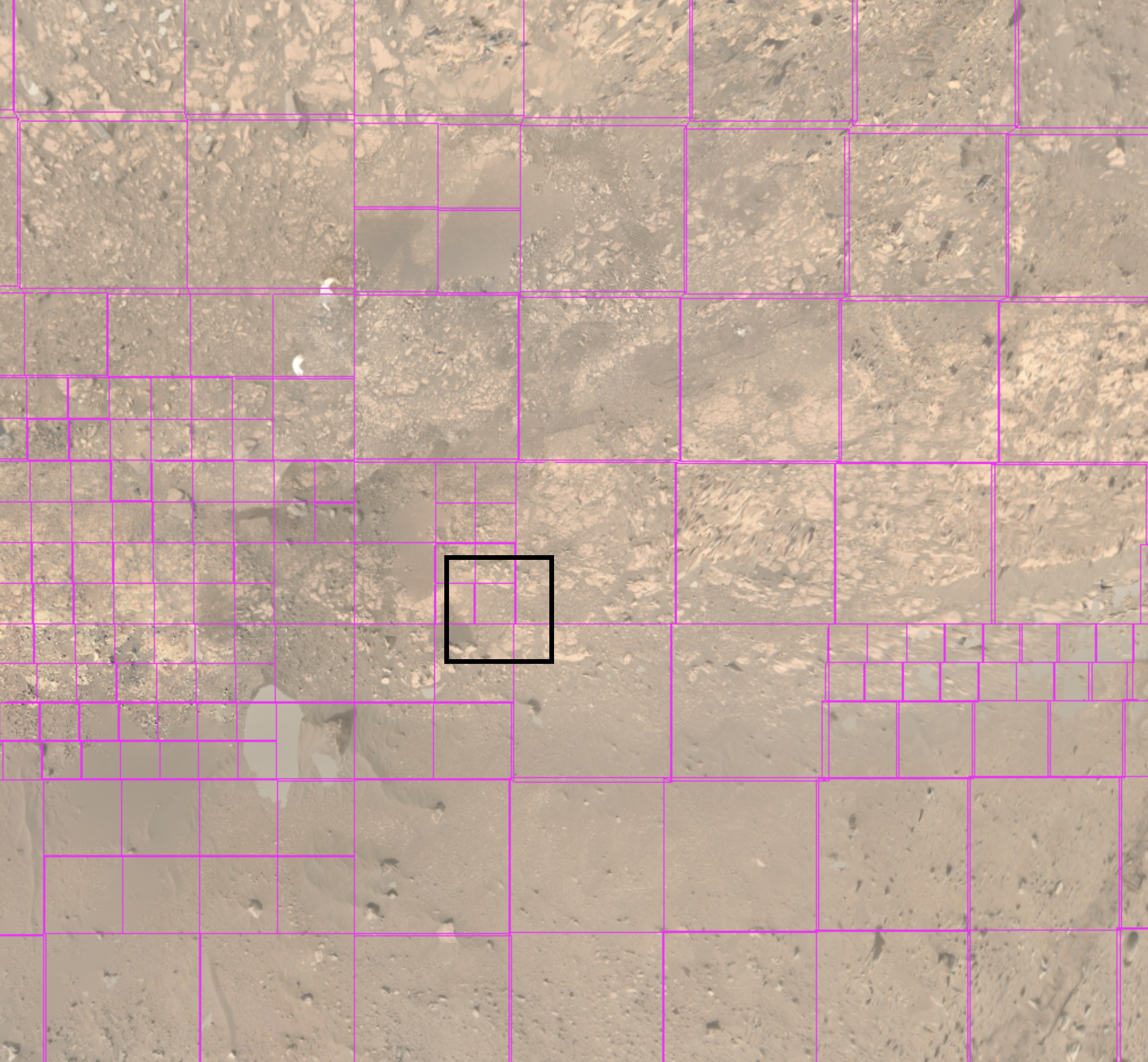}
\end{subfigure}
\hfill
\begin{subfigure}{0.33\textwidth}
    \includegraphics[width=\textwidth]{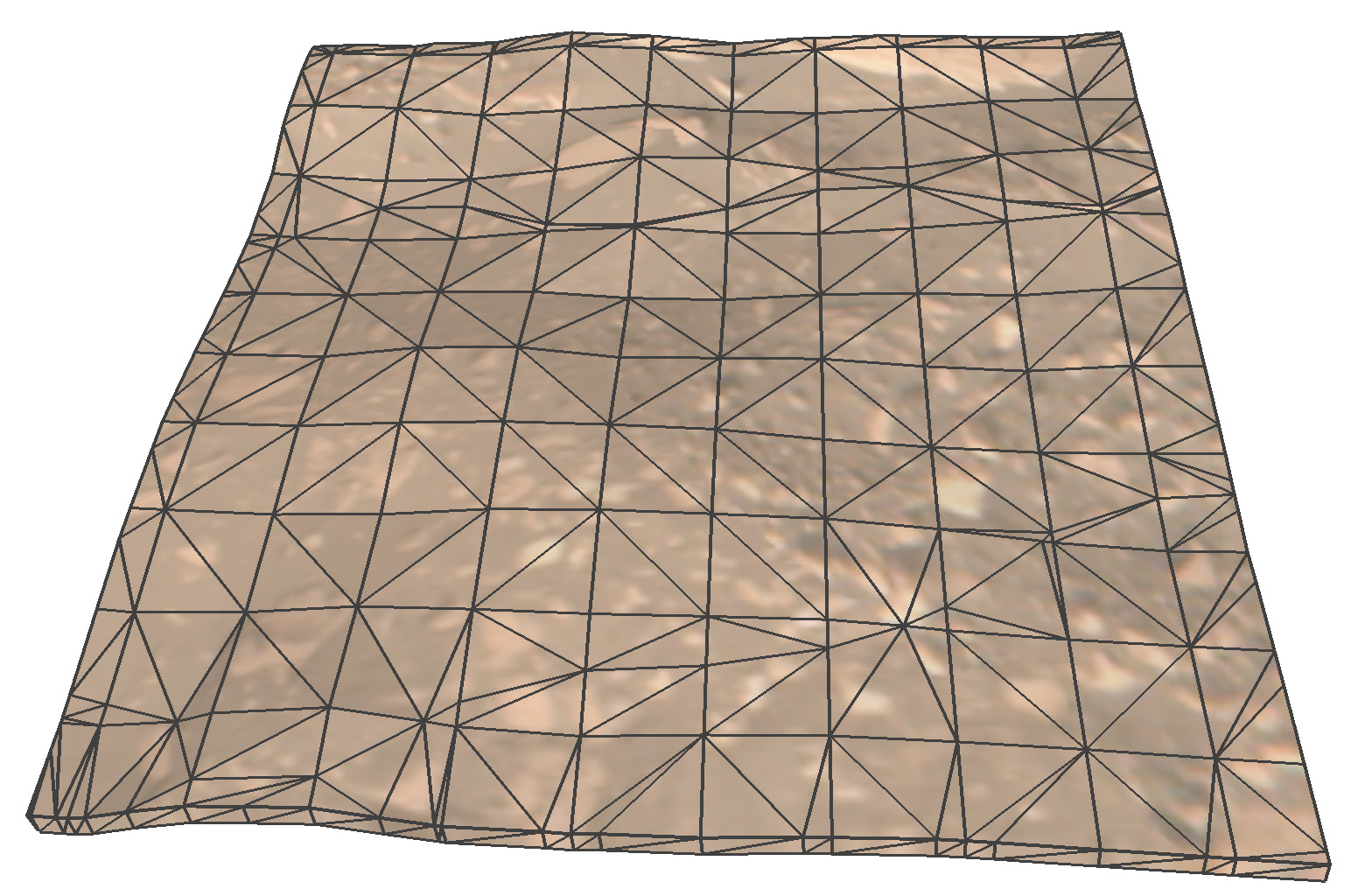}
\end{subfigure}
\hfill
\begin{subfigure}{0.32\textwidth}
    \includegraphics[width=\textwidth]{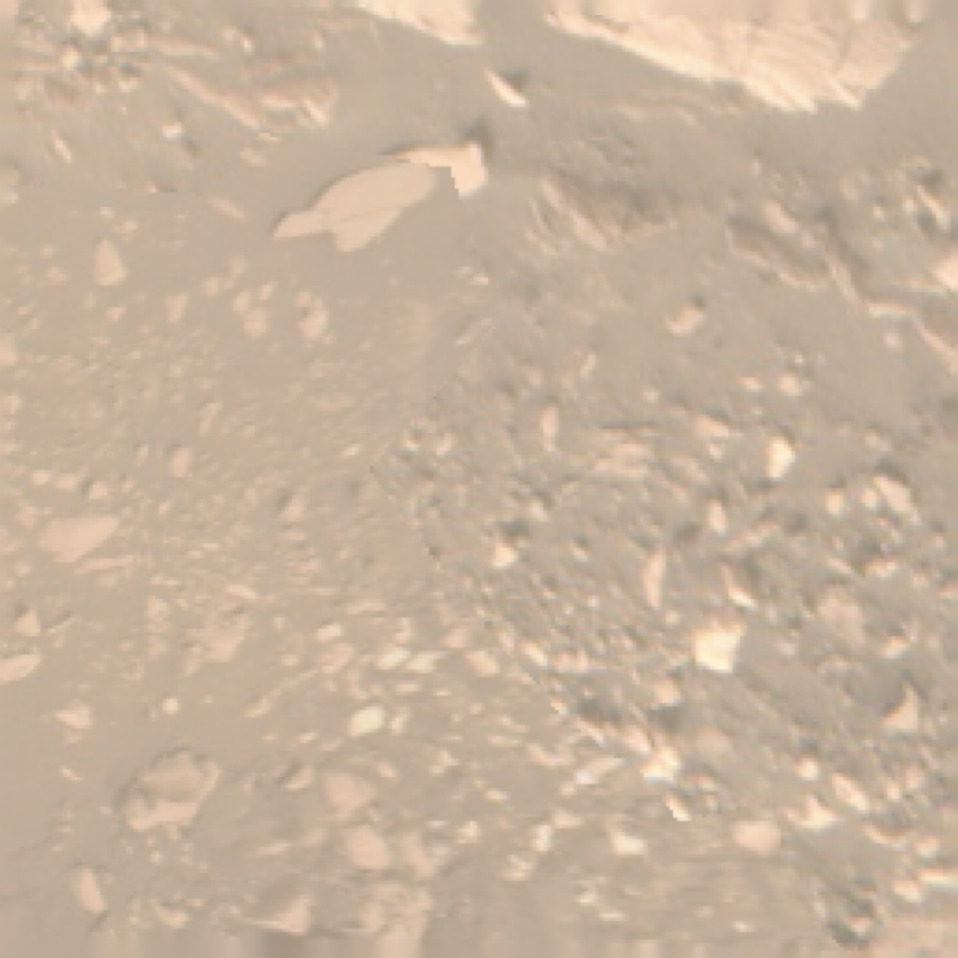}
\end{subfigure}
\caption{Example leaf tile showing the context of the leaf in the contextual mesh (left), the leaf triangle mesh with 545 triangles (center), and the 259x259 pixel leaf texture image (right).  This leaf is at depth 11 in the tile quadtree and has a 0.5x0.5x0.14m XYZ bounding box.  Part of the tile ``skirt'' is visible as the strip of narrow triangles at the bottom of the mesh.}
\label{LeafExample}
\end{figure*}

\begin{figure}
\centering
\includegraphics[width=\columnwidth]{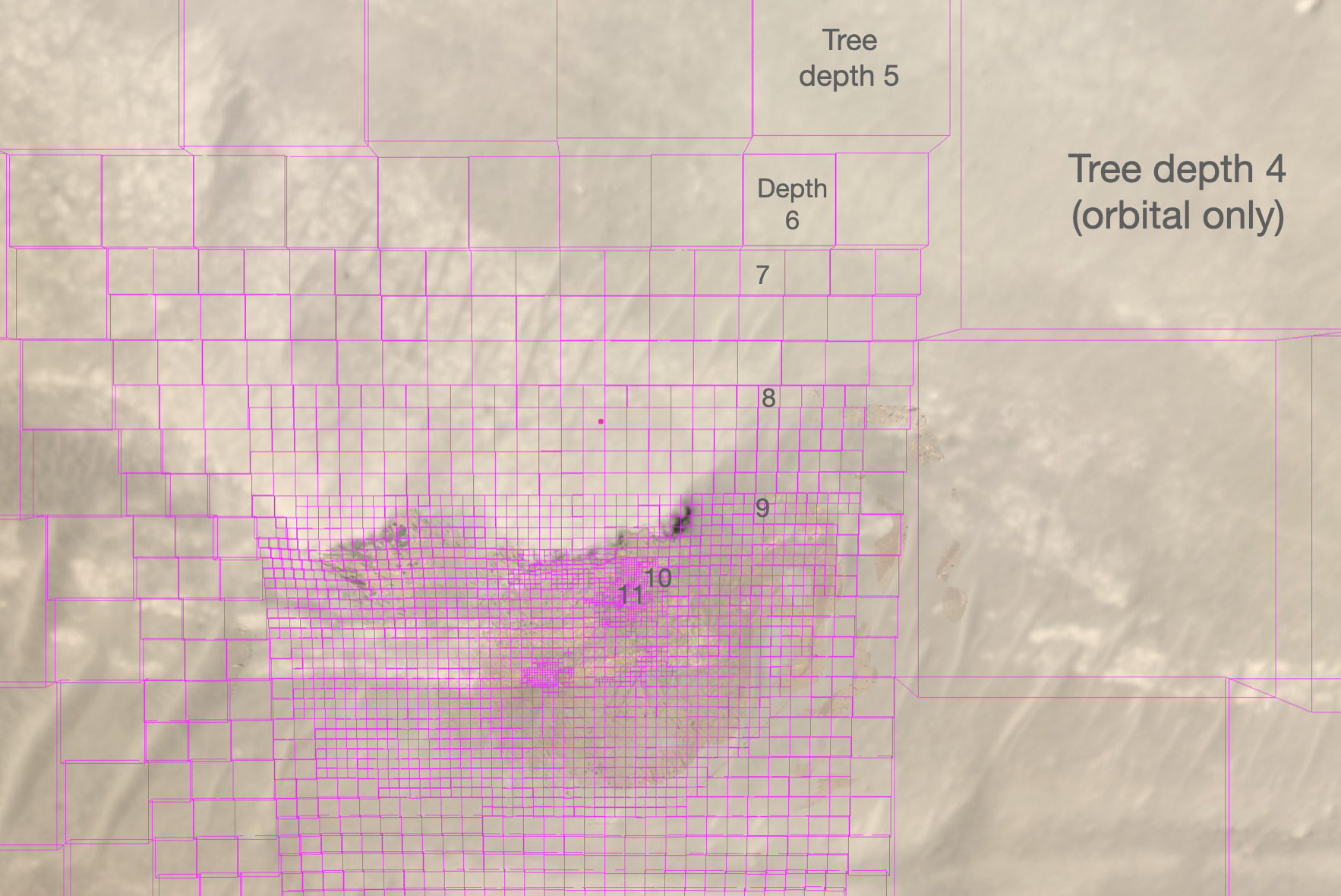}
\caption{The algorithm to define leaf tiles automatically makes the tile quadtree deeper in areas where the source data has higher resolution.}
\label{TreeDepths}
\end{figure}

Texturing will be described below.  At this point, we have the un-textured reconstructed mesh.  We now apply an algorithm to adaptively subdivide it into a quadtree of tiles.  Specifically, this quadtree defines (i) the bounds of every tile in the tree and (ii) the triangle meshes of \emph{leaf} tiles, as shown in Figures~\ref{LeafTiles} and~\ref{LeafExample}.  Importantly, the tree can have different depths along each of its branches (Figure~\ref{TreeDepths}), extending deeper---and thus subdividing into finer tiles---in areas where there is higher resolution source data.  (The figures include the texture colors which are computed later.)

\begin{figure}
\centering
\includegraphics[width=0.8\columnwidth]{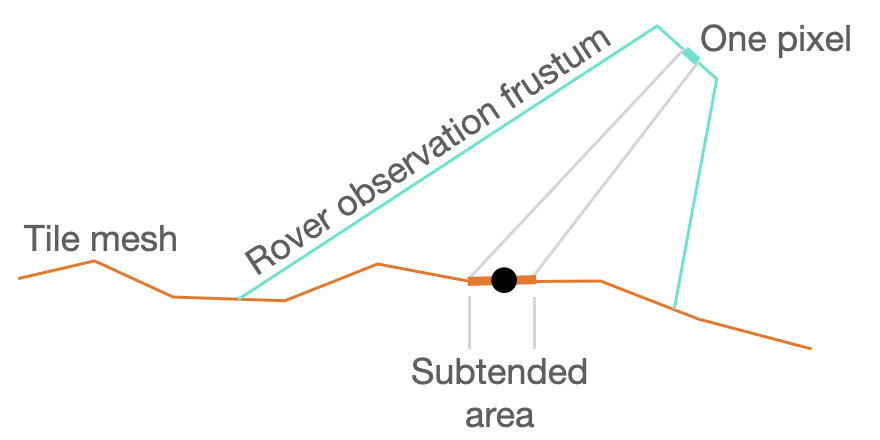}
\caption{Estimating the effective resolution of a rover observation at a 3D sample point on the terrain (black dot).  The sample point corresponds to a pixel in a nearby rover observation according to its aligned camera model.  Here only one such observation is shown, but in general there can be zero or more.  The subtended area $a_m$ in m$^2$ on the terrain mesh corresponding to the pixel that observed the point is estimated based on the camera model and the distance from the camera to the terrain at that point. This is approximately converted to an area $a_t = D a_m$ in texels$^2$ by assuming that texture stretch is relatively uniform across the tile, so there are approximately $D = w h / A$ texels per square meter where $w$ and $h$ are the tile texture resolution in texels and $A$ is the tile mesh area in m$^2$.  The effective resolution of that observation at that point on the terrain is thus approximately $1/a_t$, measured in observation pixels$^2$ per texel$^2$.}
\label{EffectiveResolution}
\end{figure}

The algorithm to compute the tile tree topology is based on the following requirements for leaf tiles:
\begin{itemize}
\item minimum horizontal extent 0.5m
\item \emph{triangle split criteria}: maximum 10,000 triangles
\item \emph{area split criteria}: maximum tile mesh area $A\leq128\text{m}^2$ ($A\leq5\text{km}^2$ for tiles that only contain orbital data\footnote{We consider a tile ``orbital only'' if it is outside the bounding box of the central portion of the mesh reconstructed from surface data.})
\item \emph{texture split criteria}: maximum 16 pixels per texel for the observation with the highest effective resolution on the tile (Figure~\ref{EffectiveResolution}).
\end{itemize}

The first two are relatively easy to compute; the third is relatively complex.  The algorithm is as follows:
\begin{enumerate}
\item Start with a root tile $p=r$ with a square bounding box\footnote{This algorithm operates primarily in the X and Y dimensions, i.e. tangent to the planetary surface.  Bounding box coordinates in the Z axis are simply computed to contain the corresponding portion of the underlying mesh.} that covers the entire reconstructed terrain.
\item Quadtree subdivide $p$ into 4 square children $a,b,c,d$.
\item Clip out the portions of the entire reconstructed triangle mesh contained in each child.
\item For each child $h\in\{a,b,c,d\}$, subdivide further (i.e. recur on step 2 with $p=h$) if and only if
  \begin{enumerate}
  \item the bounding box of $h$ has side length greater than 0.5m \emph{and}
  \item the clipped mesh for $h$ has area $A_h$ greater than 128m$^2$ (5km$^2$ for orbital-only tiles) \emph{or}
  \item the clipped mesh for $h$ contains more than 10,000 triangles \emph{or}
  \item at a sampling of points on the clipped mesh for $h$, the highest effective resolution of any rover observation image is greater than 16 observation pixels per texel.
  \end{enumerate}
\end{enumerate}
Step 4.d is where the texture split criteria must be evaluated.  To do so, we first predict the size of the texture image that we will eventually generate for $h$, by assuming that texture stretch will be approximately uniform, and computing a resolution that will afford about 1 lineal texel per 2mm (25cm for an orbital-only tile) on the clipped mesh for $h$.  Specifically the resolution is predicted as $R_h \times R_h$ where
\begin{equation}
R_h = \text{min}\left(R_{\mathrm{max}},\text{max}\left(R_{\mathrm{min}},\left\lceil\sqrt{A_h/\alpha}\right\rceil\right)\right)
\label{EqTexRes}
\end{equation}
with $R_{\mathrm{min}}=128$, $R_{\mathrm{max}}=512$, and $\alpha=625\text{cm}^2$ if $h$ is orbital-only or $\alpha=4\text{mm}^2$ otherwise.  Next, we find all rover observations $O_h$ where the 3D camera field-of-view frustum\footnote{Camera field-of-view frusta are computed from the aligned camera model intrinsics and extrinsics, and assume a far clip plane 64m away from the camera.\label{CamFrusta}} intersects the convex hull of the clipped mesh for $h$.  Then for each observation $o$ in $O_h$ we project a 5x5 grid of uniformly spaced rays from the camera onto the terrain.  We approximate the distance from the camera to the terrain as the minimum distance from the camera to the intersection points of those rays on the mesh.  We use the camera model to estimate the subtended area $a_m$ in m$^2$ of one pixel at that distance (ignoring foreshortening), and we convert that to an area in texels$^2$ as $a_t = D a_m$ where $D=R_h^2/A_h$ (again assuming uniform texture stretch).  The highest effective resolution $\rho_o$ of observation $o$ on tile $h$ in pixel$^2$ per texel$^2$ is approximated as $\rho_o=1/a_t$.  Finally, the texture split criteria indicates that the tile should be split if $$\left[\max_{o \in O_h} \rho_o\right] > \beta$$ with $\beta=16$.

Our choices for the values of $R_{\mathrm{min}}$, $R_{\mathrm{max}}$, $\alpha$, and $\beta$ are tuned to balance quality vs complexity, including both the computation required to build the contextual mesh, the number of produced tiles, and the amount of memory and bandwidth required to store and transfer them to clients.  In some cases the output tileset may not capture the full detail of particularly high-resolution surface images.  This could be adjusted, but at the cost of longer computation times and increased memory and bandwidth requirements in ASTTRO.

This process also clearly involves multiple approximations and assumptions.  We also implemented a more precise version, and we found that though it is much more expensive to compute, the final results are not significantly different in most cases.

\subsection{Tile Mesh Skirt and Texture Coordinates}
Before saving the mesh for each leaf tile we compute texture coordinates for it, and we extend its boundary with a small vertical ``skirt'' which helps hide seams between tiles, particularly when the runtime engine chooses to display tiles at different levels in the hierarchy next to each other (Figure~\ref{LeafExample}, center).  

Automatically and robustly computing texture coordinates for a large mesh with questionable topology can be very challenging.  Though the reconstructed mesh should be mostly a connected 2-manifold, it can contain topological defects, e.g. islands from Poisson reconstruction, and it can have holes and overhangs.  We reduce the difficulty by computing texture coordinates separately for each tile.  Our approach is to first attempt to assign texture coordinates assuming the tile mesh is a 2.5 dimensional heightmap relative to the average triangle surface normal, which greatly simplifies the problem.  If the area of triangles that are backfacing relative to that direction is greater than a threshold, we fall back on the UVAtlas library~\cite{UVAtlas} which implements isochart texture atlasing~\cite{Zhou04}.

In practice it typically takes roughly an hour to compute the tile tree topology, the tile bounding boxes, and to produce all the leaf tile meshes.

\section{Texturing}
\begin{figure}
\centering
\includegraphics[width=0.8\columnwidth]{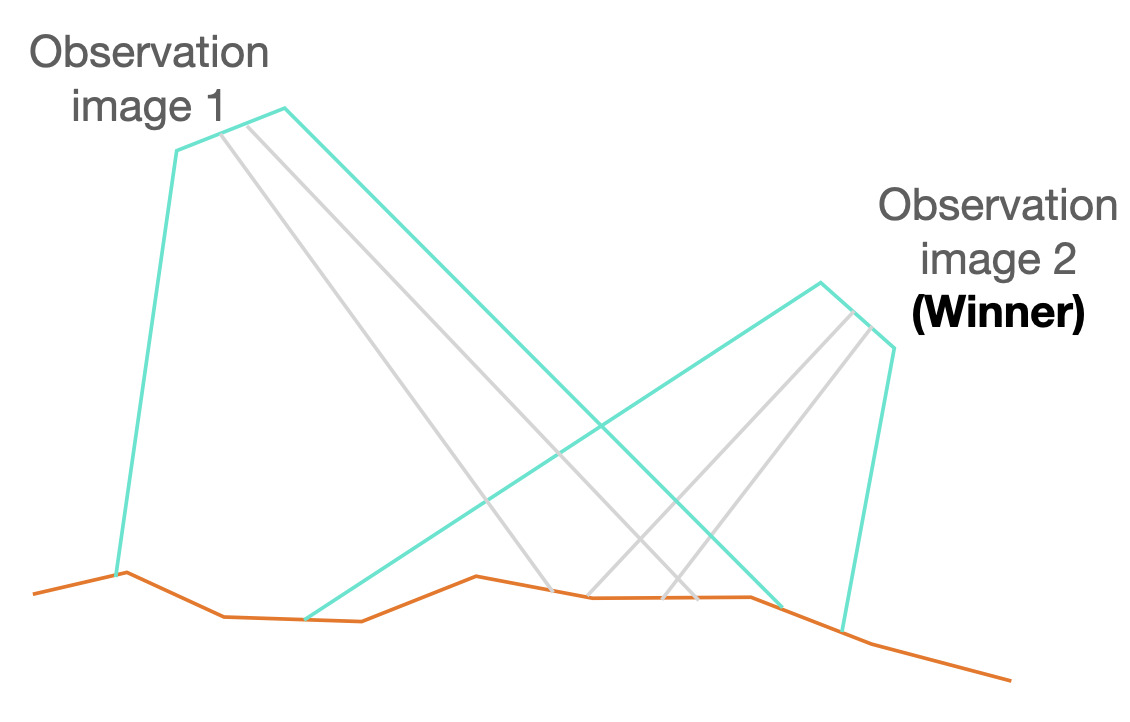}
\caption{The backprojection algorithm chooses the ``best'' observation image---typically the one with the highest effective resolution---that observed a given point on the terrain.}
\label{BackprojectSelection}
\end{figure}

\begin{figure}
\centering
\includegraphics[width=\columnwidth]{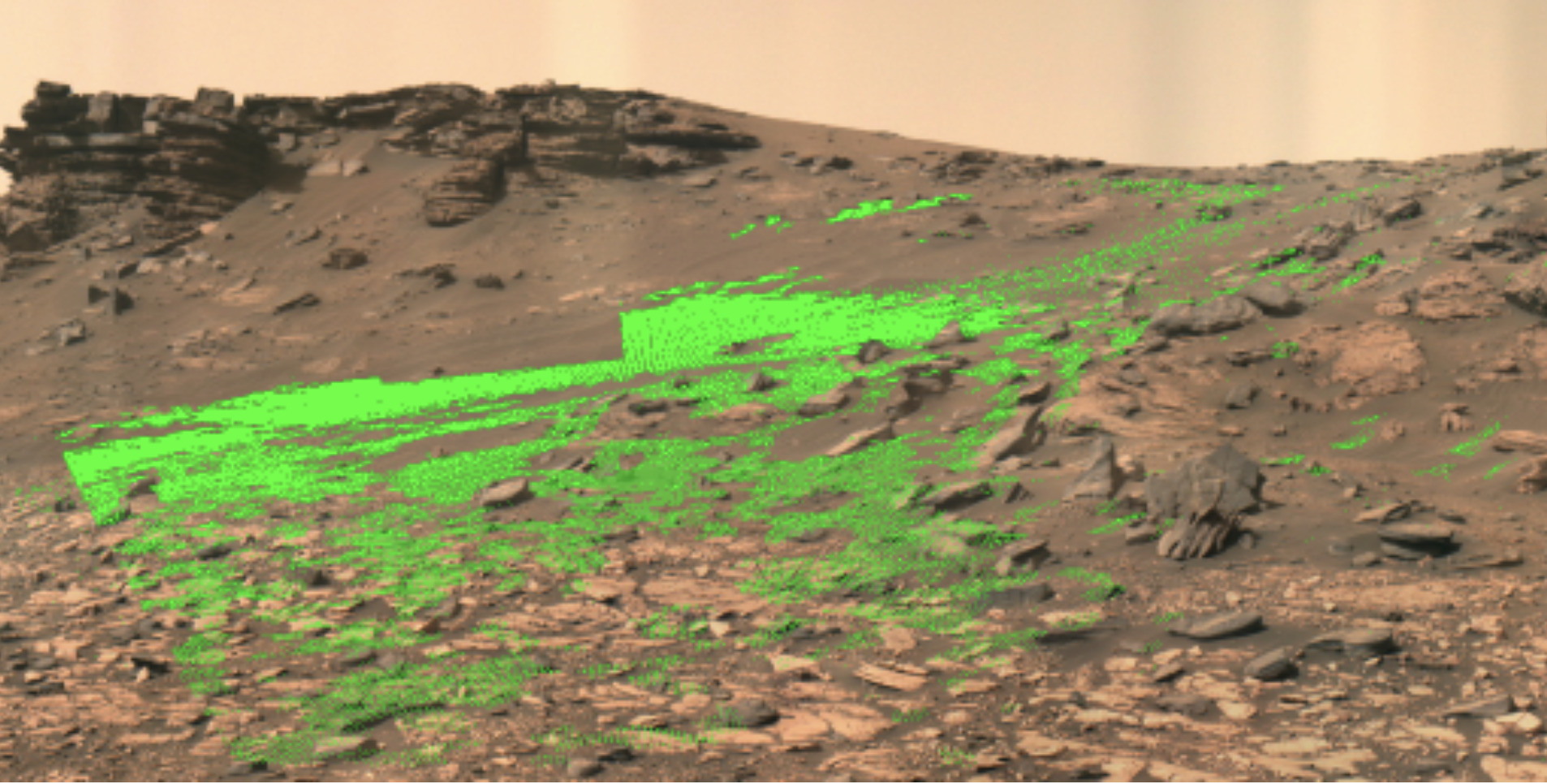}
\caption{Example rover observation image with the pixels that were selected from it by the backprojection algorithm highlighted.}
\label{BackprojectedPixels}
\end{figure}

\begin{figure*}
\centering
\begin{subfigure}{0.33\textwidth}
    \includegraphics[width=\textwidth]{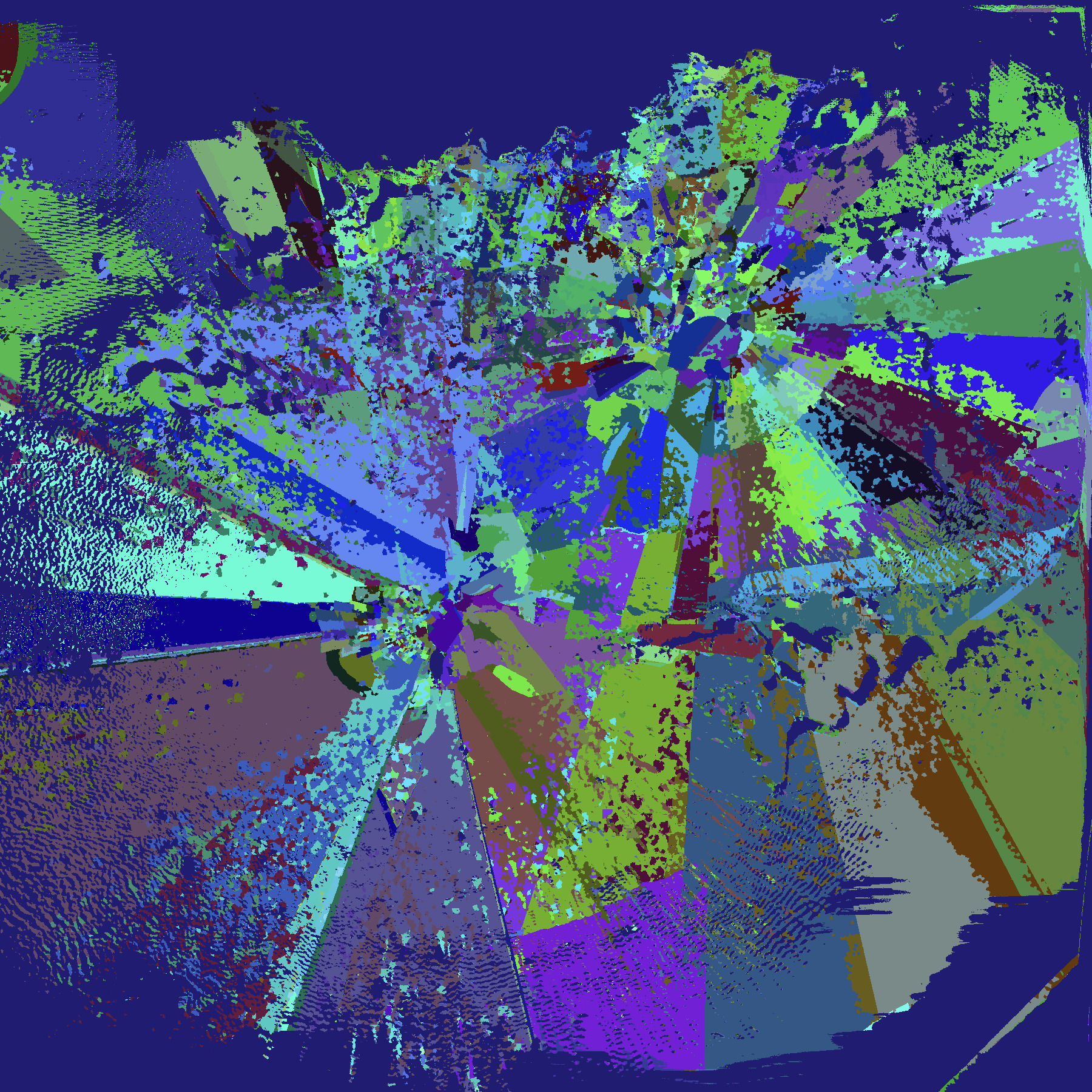}
    \caption{False color rendering showing the original observations selected by backprojection.}
    \label{SubfigBackprojectIndex}
\end{subfigure}
\hfill
\begin{subfigure}{0.33\textwidth}
    \includegraphics[width=\textwidth]{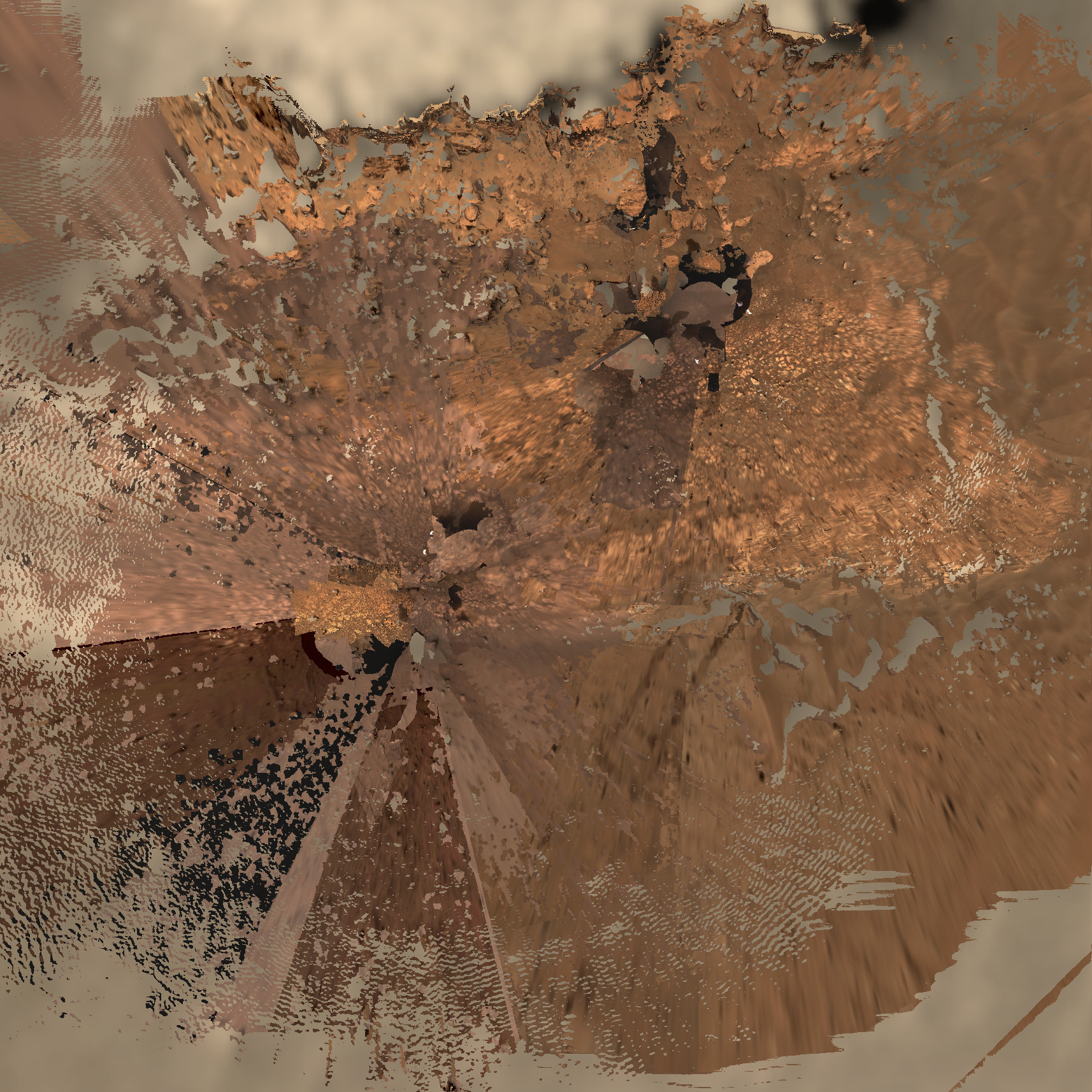}
    \caption{Raw backprojected tile textures before exposure seam correction.}
    \label{SubfigBackprojectUnblended}
\end{subfigure}
\hfill
\begin{subfigure}{0.33\textwidth}
    \includegraphics[width=\textwidth]{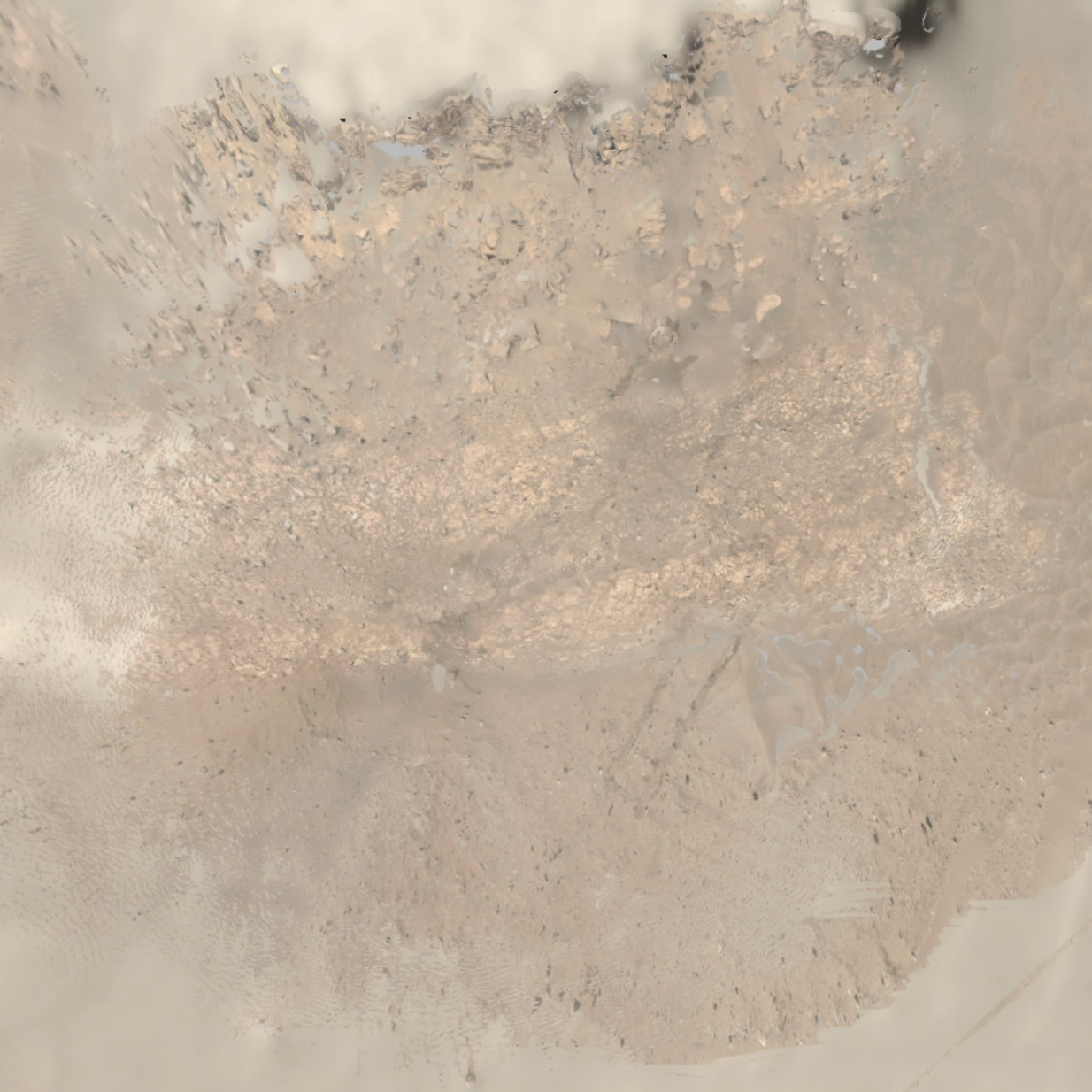}
    \caption{Backprojected tile textures after exposure seam correction.}
    \label{SubfigBackprojectBlended}
\end{subfigure}
\caption{Leaf tile textures are formed by backprojection followed by exposure seam correction.}
\label{FigBackproject}
\end{figure*}

We have now computed (i) the bounding box of every tile in the tree, and (ii) the triangle meshes of leaf tiles including texture coordinates.  The next task is to compute a texture image for each leaf tile.  Our basic approach is to backproject the color from the ``best'' observation image for each texel, preferring source images with higher effective resolution at that location (Figures~\ref{BackprojectSelection}, \ref{SubfigBackprojectIndex}, and~\ref{BackprojectedPixels}), and falling back to the orbital image where no in-situ observation is available (or if the orbital image would actually be preferable, e.g. due to foreshortening in the in-situ imagery).  This gives good resolution but typically leaves visible seams (Figure~\ref{SubfigBackprojectUnblended}), primarily due to differences in exposure brightness across the different observations.  We remove these with an exposure seam correction algorithm~\cite{KazhdanDMG} that blends brightness levels across the scene (Figure~\ref{SubfigBackprojectBlended}).

The backprojection algorithm processes all leaf tiles $h$ in parallel batches:
\begin{enumerate}
\item Compute a random sampling $S$ of points on the mesh for leaf tile $h$ at a density of 30 points per m$^2$~\cite{Corsini12}.
\item For each sample point $s\in S$ in parallel batches:
  \begin{enumerate}
  \item Find rover observation images $O_s$ that may have observed $s$ according to their aligned frusta\footref{CamFrusta}.
  \item Add the orbital image observation to $O_s$.
  \item Sort $O_s$ in decreasing order of highest effective resolution at $s$, considering the sensor resolution, instrument field of view, distance from the camera to $s$, and foreshortening\footnote{It's possible to also e.g. prefer color to grayscale images here, but we currently only do that to break ties in the other measures.}.
  \end{enumerate}
\item Choose the resolution $R_h$ of the texture image for $h$ using Equation~\ref{EqTexRes}.
\item For each texel $t$ in the $R_h \times R_h$ texture image for $h$, in parallel batches:
  \begin{enumerate}
  \item Compute the corresponding spatial point $x_t$ to $t$ on the mesh for $h$.
  \item Find the closest sample point $\min_{s_t\in S} \|s_t - x_t\|$.
  \item Attempt to backproject from $x_t$ to a pixel $p_t$ in each of the potential observing cameras in $O_{s_t}$ in priority order, but skipping any for which there is an occlusion due to either the terrain or the rover itself\footnote{We use the MXY rover mask RDR data products to determine if there was an occlusion due to the rover}.
  \item If all observations in $O_{s_t}$ were occluded (e.g. $x_t$ was in an un-observed overhang) assign a default color to the texel and abort.
  \item Otherwise take the color of the ``best'' original observation image $o_t\in O_{s_t}$ at pixel $p_t$ as the color of texel $t$.
  \end{enumerate}
\end{enumerate}
The result of this process is actually two $R_h \times R_h$ three band images for each leaf tile $h$.  One is the RGB color texture image, and the other is an \emph{index image} where
\begin{itemize}
\item Band 0 contains a unique integer that references the original observation image $o_t$ from which the color was sourced.
\item Band 1 contains the horizontal coordinate of the observation image pixel $p_t$.
\item Band 2 contains the vertical coordinate of $p_t$.
\end{itemize}
The index image will be used in the exposure seam correction algorithm below.  The index images are also saved with the contextual mesh and could potentially be leveraged to re-color it at any later time if an alternate coloring of the original observation images is desired.  For example, the M20 GDS typically computes false-color RDR products that indicate reachability and traversability~\cite{M20CamSIS}.  However, implementing such downstream re-coloring would be future work.
  
Backprojection is one of the most expensive steps in computing the contextual mesh because it involves selecting image data from among potentially thousands of observations, and this must be performed for a large number of points on the terrain to color each texel.  For the example contextual mesh 0534\_0261222 backprojection used pixels from about 1,300 rover observation images.  Our implementation is highly parallelized and typically completes in a few hours.

\subsection{Exposure Seam Correction}
The radiometric correction in the RAS rover image products compensates for the sensor response curve but not scene illumination.  We have experimented with using the RZS product type instead, which does compensate for the time of day at acquisition~\cite{M20CamSIS}.  Often this reduces but still does not eliminate exposure seams.  Also, proper RZS processing is not available in all contexts including with Earth-based test data.

\begin{figure}
\centering
\includegraphics[width=\columnwidth]{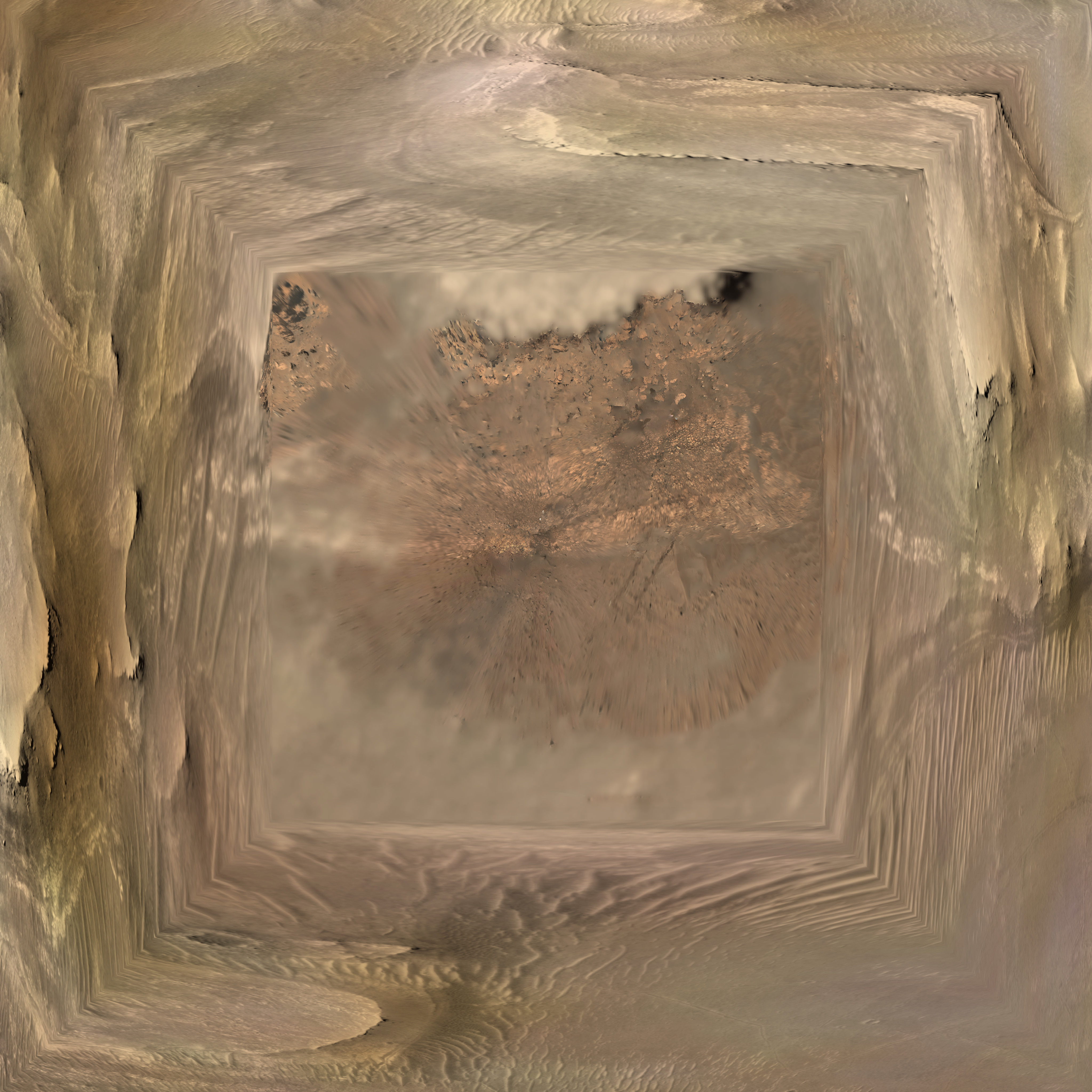}
\caption{Birds-eye view overhead rasterization at fixed resolution of $4096 \times 4096$ after computing exposure seam correction~\protect\cite{KazhdanDMG} for contextual mesh 0534\_0261222.  All leaf textures are rasterized into this image, effectively limiting and normalizing the global texture resolution.  A warping function allocates more output pixels in the central detail area and fewer in the peripheral areas which are typically textured mostly from orbital data.}
\label{BlendWarp}
\end{figure}

We have had good success instead applying a multigrid-based exposure seam correction algorithm~\cite{KazhdanDMG}, which adjusts the intensity of individual pixels to blend intensity across areas of the texture sourced from different observations.  This is fundamentally a global algorithm, but we want to avoid constructing a single ultra-high resolution global texture image.  We resolve this conflict by rasterizing all the leaf tiles into a top-down birds-eye view (BEV) at a fixed resolution of $4096 \times 4096$ and then propagate the results to corresponding areas of the original observation images.  In this case we also incorporate a nonlinear warp into the BEV rasterization so that a larger fraction of pixels in the output image are allocated to the central detail area, as shown in Figure~\ref{BlendWarp}.
\begin{figure}
\centering
\includegraphics[width=\columnwidth]{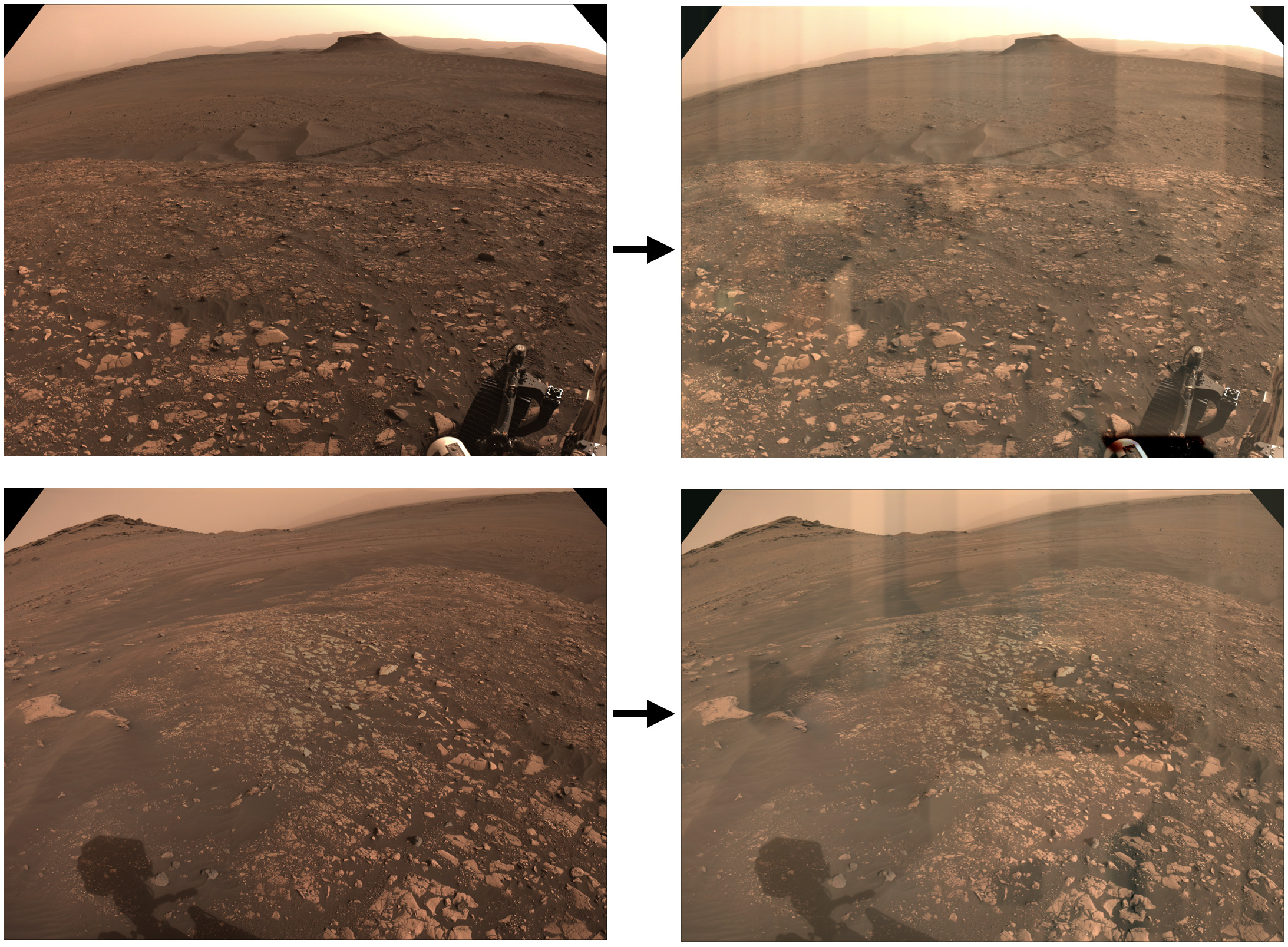}
\caption{Each original rover observation image is updated by applying the changes in brightness from the sparse set of image locations corresponding to pixels in the fixed-resolution global blend image (Figure~\ref{BlendWarp}) and then infilling those changes to remaining areas.  This figure shows two particular example observations out of the roughly 1,300 used in contextual mesh 0534\_0261222.}
\label{BlendPropagation}
\end{figure}

The algorithm to apply exposure seam correction texture areas colored from rover observation images\footnote{It should be possible to also adjust the intensity of areas colored by the orbital image, but we haven't implemented that yet.} is:
\begin{enumerate}
\item For each used original rover observation image, in parallel batches, create a copy with a Gaussian blur to suppress high frequency variations.
\item Software rasterize all leaf tiles in parallel batches into a simulated birds-eye view (BEV) from above at a fixed resolution of $4096 \times 4096$, applying a nonlinear warp in the rasterizer fragment space that allocates more of the output image to the central area, and dynamically re-coloring the leaves using the blurred observation images and the index images saved during leaf tile backprojection.  The effective output resolution ranges from about 30 pixels per meter in the central area to 2 pixels per meter in the periphery.  Two 3 band images are produced: $C$ is the RGB color image and index image $I$ holds corresponding metadata including the source observation image (band 0) and the horizontal and vertical pixel coordinates that were sampled in it (bands 1 and 2).
\item Apply the exposure seam correction algorithm from~\cite{KazhdanDMG} to $C$ and $I$, resulting in a corresponding $4096 \times 4096$ blended image $B$ (Figure~\ref{BlendWarp}).
\item For each rover observation image index $i$ occurring in $I$ in parallel batches:
  \begin{enumerate}
  \item For each pixel $(x,y)$ in $I$ where $I(x,y,0)=i$ (i.e. the value of band 0 at $(x,y)$ in $I$ is observation index $i$) apply the difference $\text{intensity}(B(x,y)) - \text{intensity}(C(x,y))$ to observation image $i$ at pixel $(I(x,y,1), I(x,y,2))$.
  \item Infill the intensity changes to all other pixels of observation image $i$ (Figure~\ref{BlendPropagation}).
  \end{enumerate}
\item For each leaf tile $h$ in parallel batches, recompute the tile texture using the saved index image for $h$ and the new blended observation images.
\end{enumerate}

Exposure seam correction typically takes about an hour. If some of the input images are in color and others are not, a median hue can optionally be computed at this stage and applied to colorize the grayscale data.  Currently the mission operations use-case prefers to keep the original colorspace, so colorization is disabled by default (e.g. Figure~\ref{SubfigContextual} shows areas with both color and grayscale texture data).

\section{Hierarchical Tiling}
\begin{figure*}
\centering
\begin{subfigure}{0.245\textwidth}
    \includegraphics[width=\textwidth]{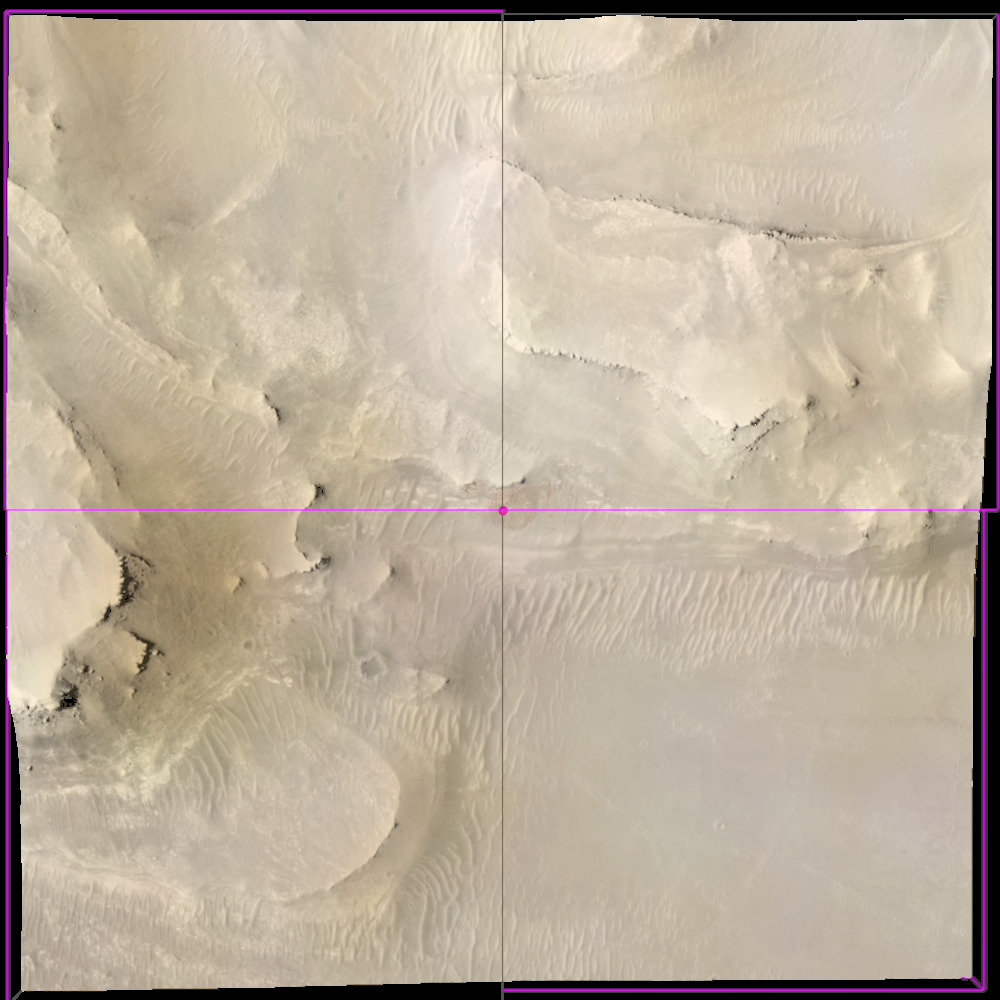}
\end{subfigure}
\hfill
\begin{subfigure}{0.245\textwidth}
    \includegraphics[width=\textwidth]{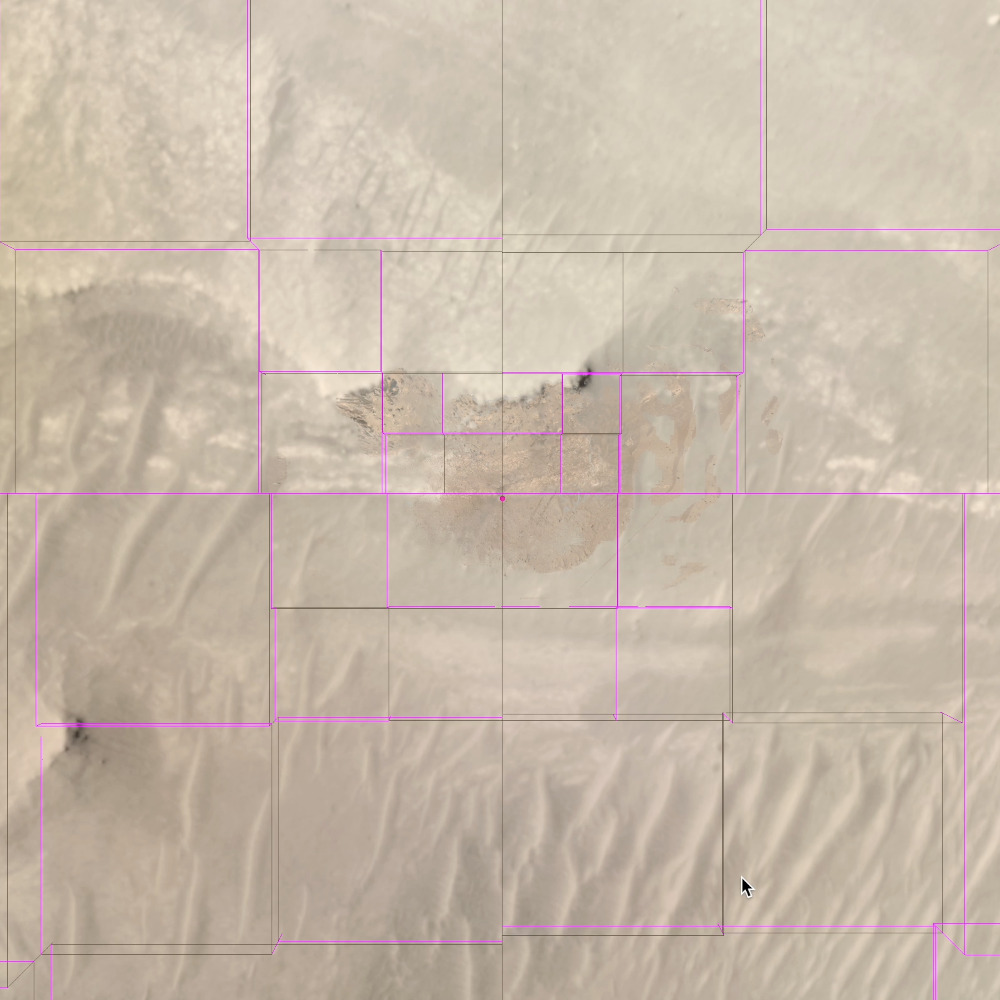}
\end{subfigure}
\hfill
\begin{subfigure}{0.245\textwidth}
    \includegraphics[width=\textwidth]{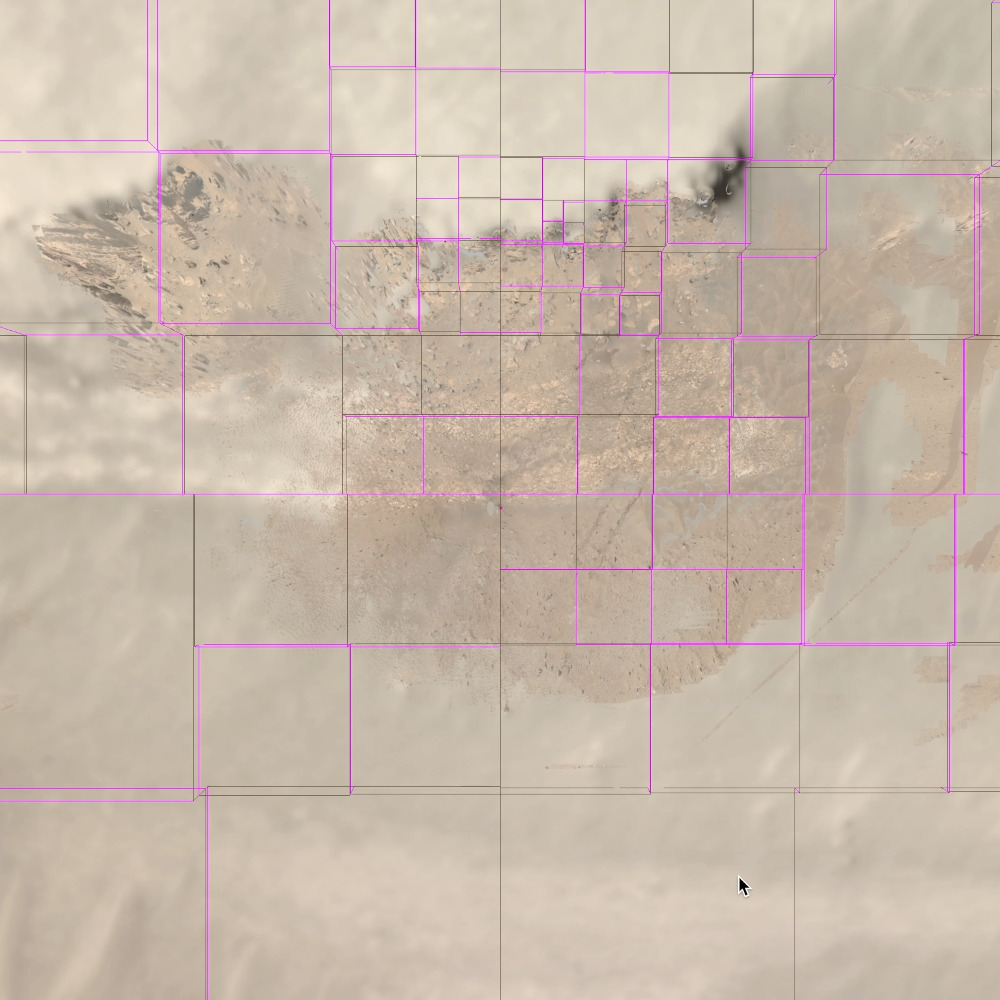}
\end{subfigure}
\hfill
\begin{subfigure}{0.245\textwidth}
    \includegraphics[width=\textwidth]{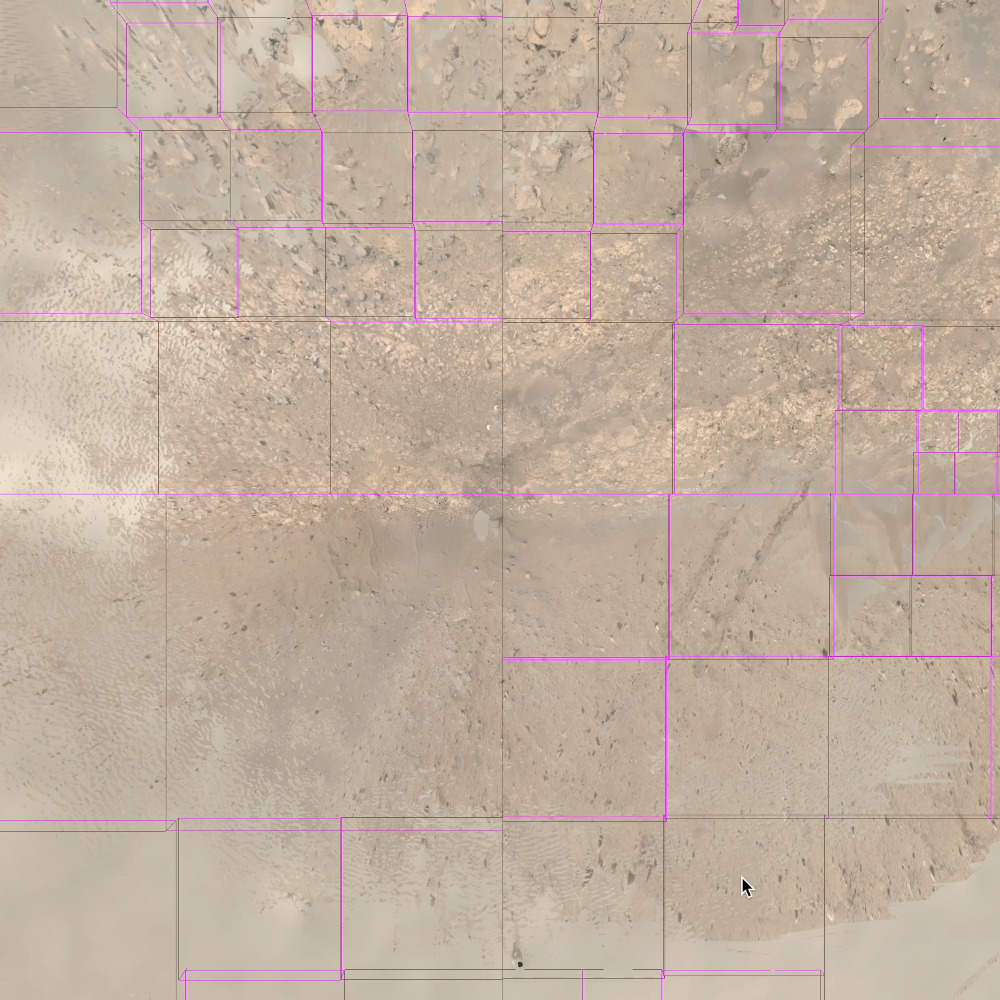}
\end{subfigure}
\caption{Zooming in, from left to right, starting from the full 1km square extent of the contextual tileset 0534\_0261222, showing the parent tiles that are active in each case.  Parent tiles correspond to the non-leaf nodes in the tile quadtree.  They approximate the geometry and texture of their children, and are used automatically by the viewer when the additional detail of the finer tiles wouldn't be visible at the current distance.}
\label{ParentTiles}
\end{figure*}

We have now computed (i) the bounding box of every tile in the tree, (ii) the triangle meshes of leaf tiles, and (iii) the textures of leaf tiles.  To complete the terrain tileset the last step is to compute the triangle meshes and textures of non-leaf \emph{parent} tiles.  These represent the mesh at coarser levels of detail, and are the key to enabling ASTTRO to quickly load and progressively refine its rendering as the user navigates, as shown in Figure~\ref{ParentTiles}.

Parent tile meshes and textures are built bottom-up by combining and then coarsening the meshes and textures of spatially overlapping tiles from the next-finer level.  Typically these are the topological children of the tile, however, we also include adjacent ``cousins'' in the computation to avoid boundary effects when creating the parent tile mesh.

For each parent tile $p$ a geometric error metric $\epsilon^g_p$ is also computed as the Hausdorff distance~\cite{Guthe05} in meters between its mesh and the meshes of the finer tiles $F_p$ that it was built from.  A texture error metric $\epsilon^t_p$ is also estimated as the expected length in meters on the mesh subtended by 4 texels.  The tile error is defined as the maximum of these two quantities plus the maximum error in $F_p$: $$\epsilon_p = \text{max}(\epsilon^g_p, \epsilon^t_p) + \max_{h \in F_p} \epsilon_h$$ where leaf tiles $l$ have $\epsilon_l=0$.

\begin{figure}
\centering
\includegraphics[width=\columnwidth]{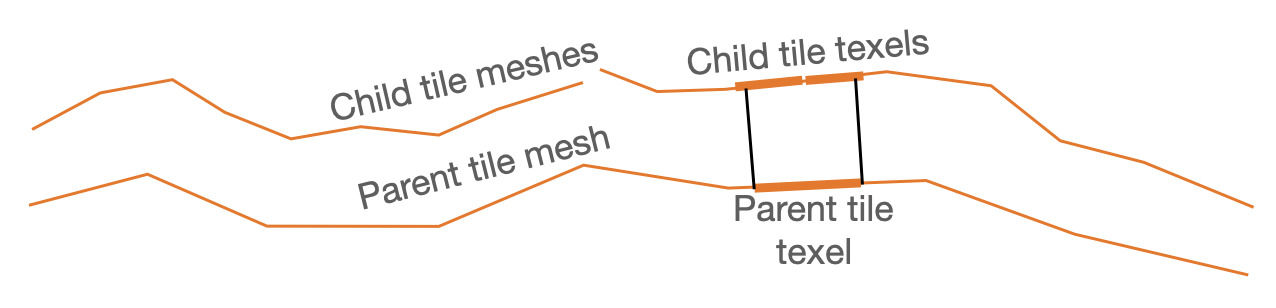}
\caption{Texel colors are ``baked'' to a parent tile from those of its children.  For every texel on the parent tile we sample a texture color from a corresponding spatial point on the children.}
\label{TextureBaking}
\end{figure}

The full algorithm to compute parent tile meshes and textures is:
\begin{enumerate}
\item Maintain a queue $Q$ of parent tiles $p$ for which all the finer tiles $F_p$ required to build $p$ are already completed.  $Q$ initially contains all parents of leaf tiles.
\item While $Q$ is not empty, extract a set of parents from $Q$ and build them in parallel.  For each parent $p$:
  \begin{enumerate}
  \item Union the meshes of all tiles in $F_p$ to create an initial parent tile mesh $m_0$.
  \item If $m_0$ has less than 10,000 triangles, use it as the parent tile mesh $m_p=m_0$.  Otherwise:
    \begin{enumerate}
      \item Randomly resample~\cite{Corsini12} $m_0$ as a pointcloud $p_1$ with approximately 1.5 samples for every triangle in $m_0$.
      \item Reconstruct a lower resolution mesh $m_1$ from $p_1$ using floating scale surface reconstruction (FSSR)~\cite{FuhrmannFSSR}.
      \item If $m_1$ still has more than 10,000 triangles apply quadric edge collapse~\cite{Garland97} to simplify it to that threshold.
      \item Use $m_p=m_1$ as the parent tile mesh.
    \end{enumerate}
  \item Clip $m_p$ to the parent tile bounding box.
  \item Determine the resolution of the parent tile texture image using Equation~\ref{EqTexRes}.
  \item Compute texture coordinates for $m_p$ in the same way as was described above for leaf tiles.
  \item ``Bake'' the colors of every texel on the parent tile by sampling the colors at the corresponding locations on the tiles in $F_p$ (see Figure~\ref{TextureBaking}).
  \item Compute the error metric $\epsilon_p$ as described above.
  \item Add a skirt to $m_p$ in the same way as was described above for leaf tiles.
  \end{enumerate}
\end{enumerate}

It typically takes up to several hours to produce meshes, textures, and error metrics for all parent tiles, typically computing many at once in parallel.

\subsection{Runtime Tile Selection}
\begin{figure}
\centering
\begin{subfigure}{\columnwidth}
  \includegraphics[width=\textwidth]{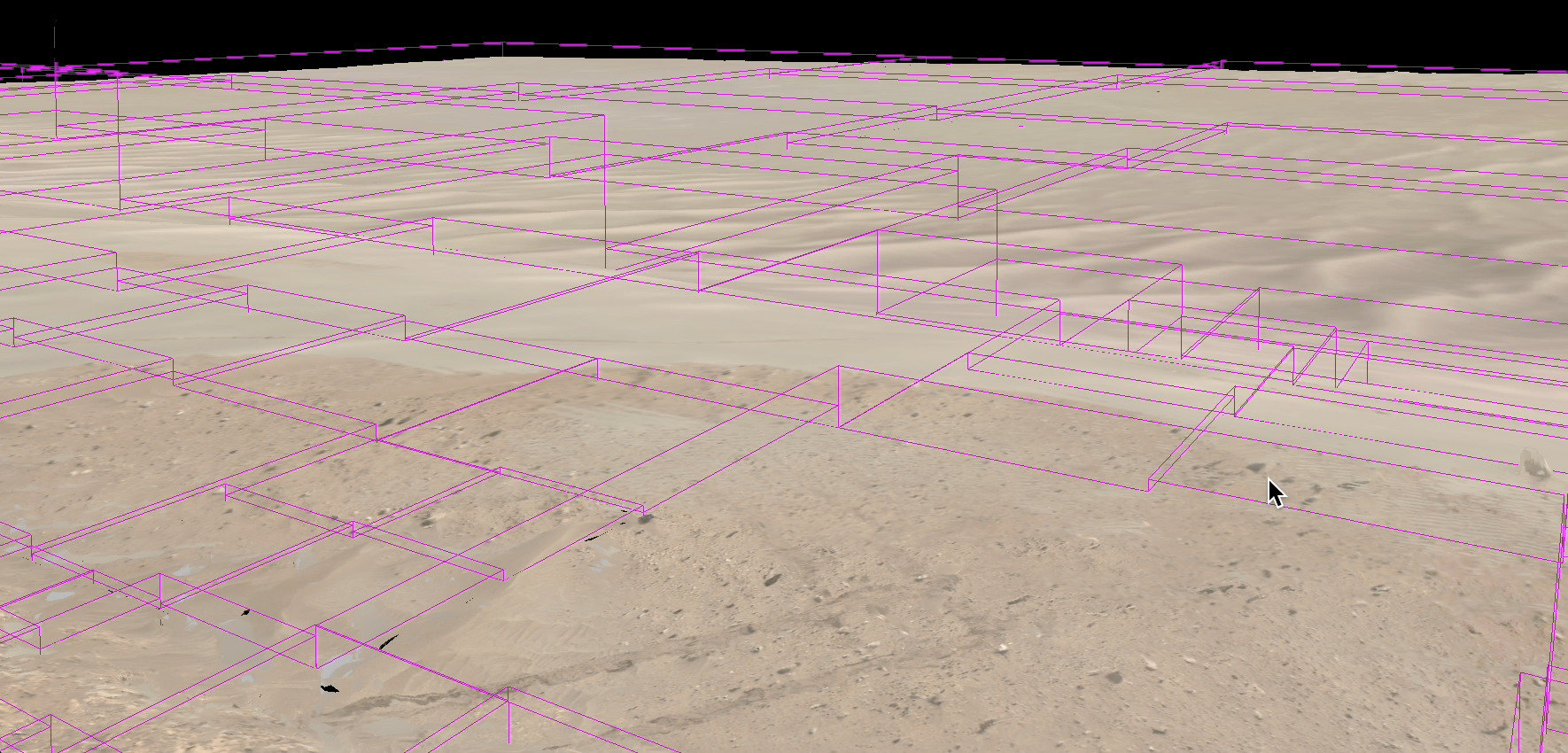}
\end{subfigure}
\begin{subfigure}{\columnwidth}
  \includegraphics[width=\textwidth]{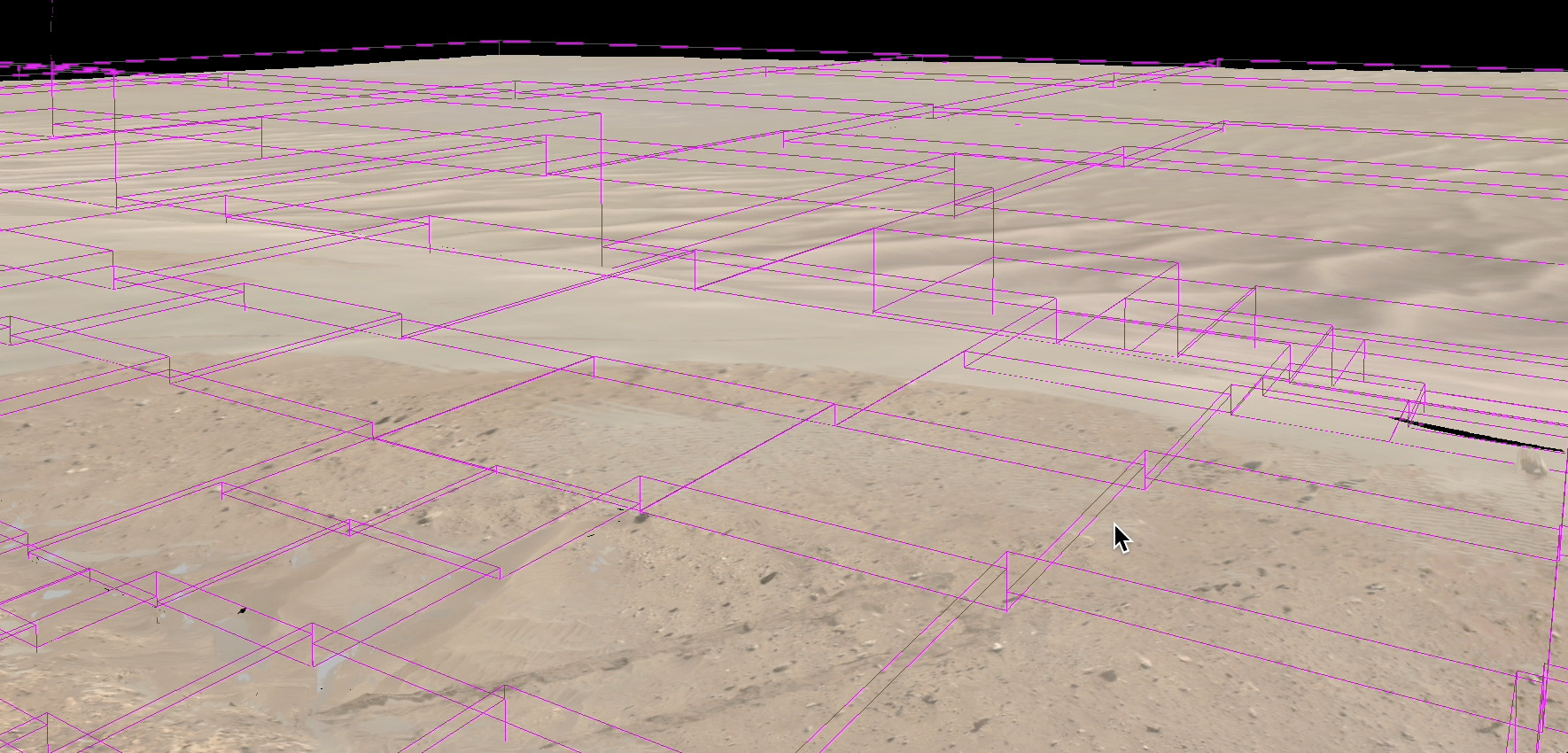}
\end{subfigure}
\begin{subfigure}{\columnwidth}
  \includegraphics[width=\textwidth]{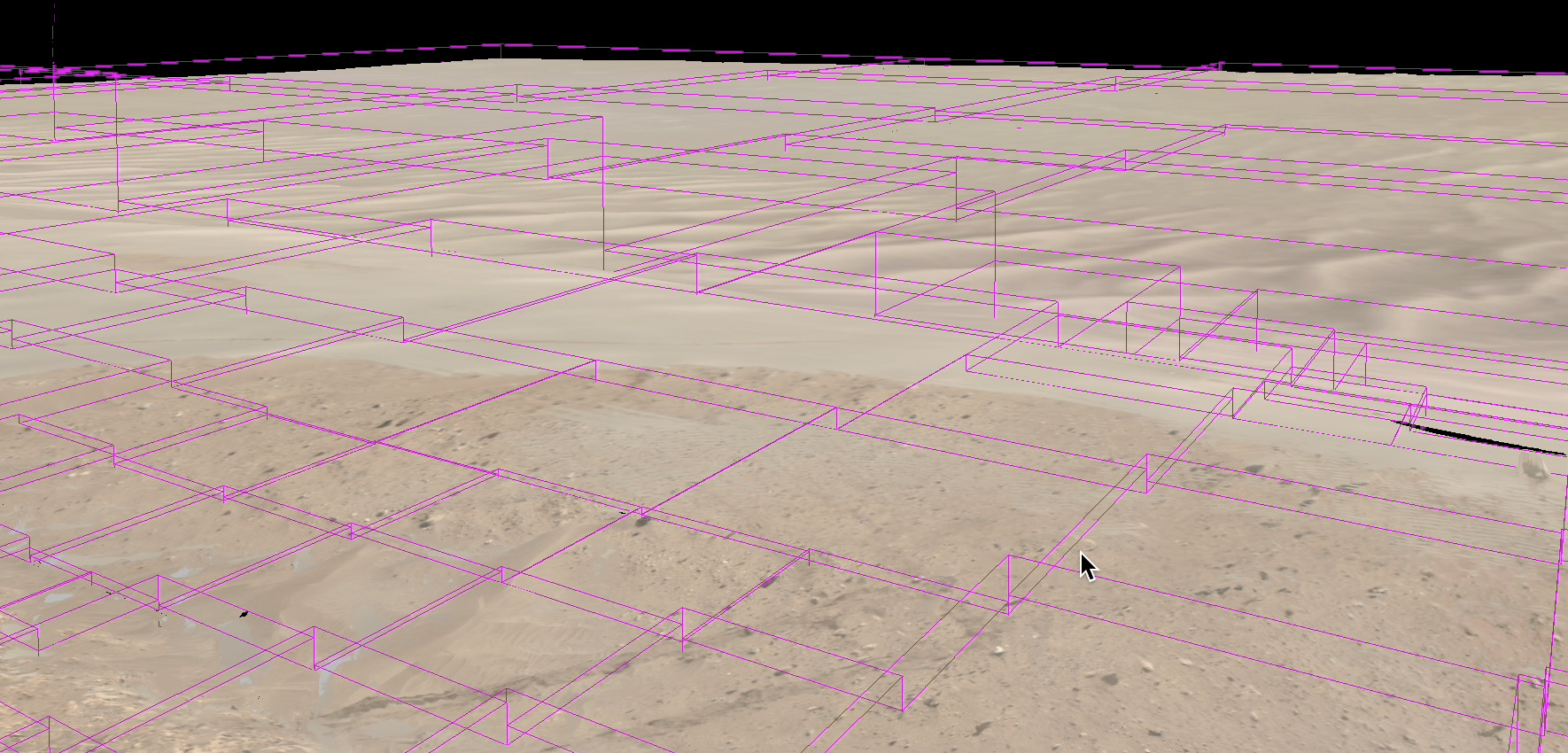}
\end{subfigure}
\caption{Example of dynamic refinement in the runtime engine as tiles are downloaded.  At top the viewer has just navigated to a new perspective and only some coarser resolution tiles are available on the client.  After a short time, as shown in the middle and bottom, finer tiles which better meet the screenspace error metric are downloaded.}
\label{TileRefinement}
\end{figure}

The rendering engine in ASTTRO implements a dynamic tile selection algorithm based on a maximum screen space error $\delta$, typically 4 pixels.  As the user navigates in the scene, candidate tiles are tracked by finding all tiles $C$ with bounds intersecting the current viewer camera frustum.  For each tile $c \in C$ the tile error $\epsilon_c$ is transformed from meters on the tile to pixels on the screen $\epsilon^s_c$ based on the current distance from the viewer to the tile, the field of view angle, and the screen resolution.  The engine then attempts to render the coarsest tiles in $C$ where $\epsilon^s_c < \delta$.  A memory limited cache of tiles is maintained on the client, and tiles are downloaded as needed.  Nearest available parent or ancestor tiles are rendered while downloads are in progress.  Figure~\ref{TileRefinement} shows an example where the runtime engine swaps in finer tiles in the foreground as they become available.

\section{Sky Sphere Tileset}
The tileset built in the steps above models the terrain typically within a 1km square around the central rover position, but not beyond.  Often there are terrain features such as hills, ridges, and mountains that would be visible on the horizon at greater distances.  Also, of course, there is no representation of the atmosphere itself.  To capture these we build a second \emph{sky sphere} tileset.  ASTTRO renders both tilesets---terrain and sky sphere---simultaneously when the contextual mesh is enabled.

\begin{figure*}
\centering
\begin{subfigure}{0.273\textwidth}
    \includegraphics[width=\textwidth]{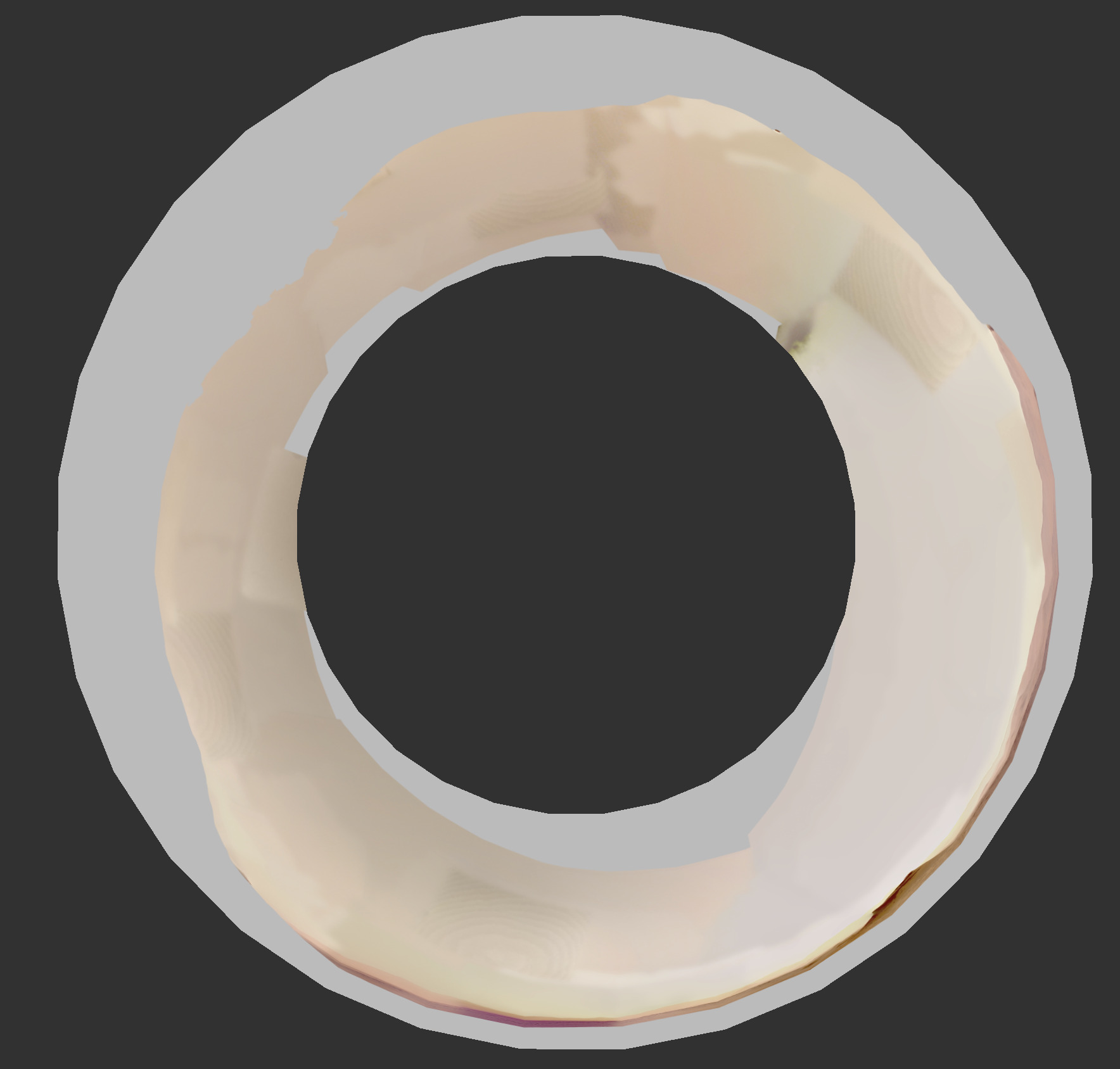}
    \caption{Sky tileset overhead view.}
\end{subfigure}
\hfill
\begin{subfigure}{0.72\textwidth}
    \includegraphics[width=\textwidth]{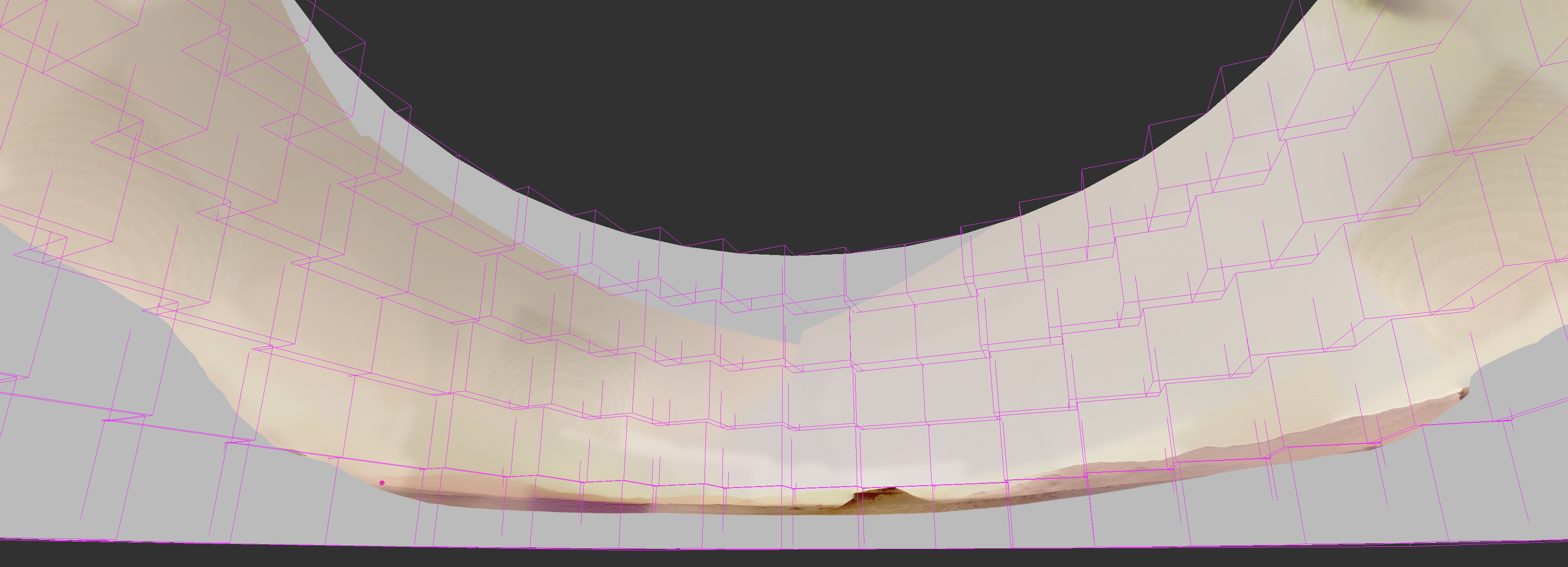}
    \caption{Sky tileset showing individual leaf tiles.}
\end{subfigure}
\caption{Textured sky sphere tileset for contextual mesh 0534\_0261222.  The sphere diameter is about 1.6km, but the texture depicts terrain and atmospheric features that are potentially much further away.}
\label{FigSky}
\end{figure*}

The sky tileset is based on a synthetic triangle mesh that tesselates the corresponding spherical surface.  The sphere radius is automatically chosen to be a bit larger than the diagonal length of the terrain tileset, so for a typical 1km square terrain the sky sphere is about 1.6km in diameter.  Leaf tiles are cut from the mesh in a similar manner as was described above for the terrain tileset.  Only an annular portion of the sphere is tiled, extending from about 10 degrees below to about 40 degrees above the horizon, as shown in Figure~\ref{FigSky}.  Unlike the terrain tileset, for the sky sphere only leaf tiles are created without a hierarchy of parent tiles at coarser levels of detail.  Each sky sphere leaf tile spans about 10 degrees horizontally and vertically (also shown in the figure).  We have found that to be coarse enough to avoid the need for parent tiles, but fine enough to support efficient dynamic tile streaming depending on the user's viewpoint at runtime.

The sky tileset is textured by backprojecting all rover image observations onto the sky sphere using essentially the same algorithm as described above for texturing the terrain itself.  One key difference is that here any point on the sky sphere which is occluded by terrain is masked (gray in Figure~\ref{FigSky}).  The sky sphere tile textures are then exposure seam corrected using the same algorithm as described above for the terrain, but with a slight modification to accommodate the circular topology of the sky sphere tileset.

At a high level the sky tileset is closely related to well-known approaches to synthesize and view 360 degree environment map image panoramas~\cite{SzeliskiPano,GreeneEnvMap,ChenQTVR}, but reformulated to use the same tiling, backprojection, and exposure seam correction algorithms as for the main terrain tileset.

It typically takes only a few minutes to build the sky sphere and it adds only a moderate amount of runtime overhead when viewing in ASTTRO.  However, when using the contextual mesh, the sky sphere often seems to have a major contribution to spatial awareness---it is typically the only way that major landmarks, such as distant mountains and ridges, can be seen.

\section{Summary and Future Work}
Landform has now produced over 2000 contextual meshes since the landing of Perseverance on Mars in February of 2021.  Most of these are only available for internal use by the mission team, however, a subset have been made available for interactive public viewing on the Explore with Perseverance~\cite{ExploreWithPerseverance} website.  We also recently received approval to release the entire Landform codebase to the public under an open source license, which should be complete by the time of publication.  While the ASTTRO application is not open source, we have already released the Unity3DTiles engine it uses to load and render contextual meshes~\cite{Unity3DTilesSource}.  And our colleague Garrett Johson has also released a compatible pure javascript runtime engine~\cite{3DTilesJSSource}.

We look forward to involve the open source community in further development.  In particular, some areas of future work include
\begin{itemize}
\item {\bf PDS archive}---At present Landform is only set up to interface directly to mission systems (e.g. IDS pipeline, ODS, PLACES) that are not accessible by the general public.  However, all the relevant data is also periodically delivered to the planetary data system archive~\cite{M20PDS}, so it should be possible to adapt Landform to work from that instead.  Also, the Landform contextual mesh is not currently saved back to PDS, but with some effort it may be possible to produce PDS compatible artifacts.
\item {\bf Performance}---The current build time of 4--8 hours is often longer than desired.  Though we have worked hard to optimize the implementation, there may still be room to make improvements.
\item {\bf Full pose graph optimization}---One rough edge in the current implementation is that we don't handle loop closures in the pose graph of the BEV aligner.
\item {\bf Orbital blending}---Another rough edge in the current implementation is that we don't apply exposure seam correction to areas textured by the orbital image.  This should be easily fixed.
\item {\bf Overlays} The texture index images saved with each contextual mesh and could potentially be leveraged to re-color it, but so far this hasn't been utilized.  For example, the M20 GDS typically computes false-color RDR products that indicate reachability and traversability~\cite{M20CamSIS}.
\item {\bf Arm, helicopter, and EDL cameras}---It would be exciting to use data from additional cameras in the contextual mesh, including the SHERLOC-WATSON and PIXL-MCC cameras on the rover arm, the Ingenuity helicopter cameras, and the entry, descent, and landing cameras.  These all present challenges to incorporate in the contextual mesh, but are well within the realm of possibility using modern techniques.
\end{itemize}

\acknowledgments
The work described in this paper was carried out at the Jet Propulsion Laboratory, California Institute of Technology, under a contract with the National Aeronautics and Space Administration.

The development and implementation of the Landform contextual mesh has been a joint effort of many contributors including Alex Menzies, Charles Goddard, Bob Crocco, Thomas Schibler, and Jeff Norris.  It is based on approaches and lessons learned from prior work on the OnSight system for the Mars Science Laboratory (MSL) mission~\cite{Abercrombie17}.

\bibliographystyle{IEEEtran}
\bibliography{Vona__2025__Landform_Contextual}

\thebiography
\begin{biographywithpic}{Marsette Vona}{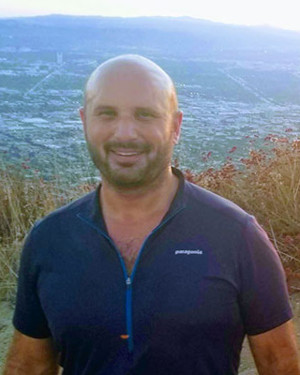} develops user interface software at JPL with a focus on robotic surface missions. He helped build the Science Activity Planner ground software for the Mars Exploration Rover mission and is currently lead developer for several Mars 2020 ground software subsystems including the Landform service for terrain reconstruction.
\\
Vona was on the faculty of the College of Computer Science at Northeastern University from 2010-14, where he taught graphics, robotics, and introductory computer science and founded a robotics research group. Vona was the recipient of the NASA Software of the Year award in 2004 and the National Science Foundation CAREER award in 2012.  He received his Ph.D. in electrical engineering and computer science from MIT in 2009, his M.S. in EECS from MIT in 2001, and his B.A. from Dartmouth College in 1999.
\end{biographywithpic} 

\end{document}